\pdfoutput=1

\documentclass[11pt]{article}

\usepackage[preprint]{acl}
\usepackage{hyperref}

\usepackage{times}
\usepackage{latexsym}

\usepackage[T1]{fontenc}

\usepackage[utf8]{inputenc}

\usepackage{microtype}

\usepackage{inconsolata}

\usepackage{graphicx}

\usepackage{url}            %
\usepackage{booktabs}       %
\usepackage{amsfonts}       %
\usepackage{nicefrac}       %
\usepackage{microtype}      %
\usepackage{amsmath}
\usepackage{graphicx}
\usepackage{subcaption}
\usepackage{adjustbox}
\usepackage{siunitx}
\usepackage{multirow}
\usepackage{rotating}
\usepackage{cuted}
\usepackage{xcolor}
\usepackage{tikz}
\usepackage{comment}
\usepackage{enumitem}
\usepackage{amsmath}
\usepackage{soul}
\usepackage{pifont}
\usepackage{fontawesome}
\usepackage{colortbl}
\usepackage{tabularx}
\usepackage{tocloft}
\usepackage{etoolbox}
\usepackage{blindtext}

\newcommand{\adjustimg}{%
  \hspace*{\dimexpr\evensidemargin-\oddsidemargin}%
}

\newcommand{\centerimg}[2][width=\textwidth]{%
  \makebox[\textwidth]{\adjustimg\includegraphics[#1]{#2}}%
}

\newcommand{\loco}[0]{\textsc{LoCoVQA}}

\title{Losing Visual Needles in Image Haystacks: Vision Language Models are Easily Distracted in Short and Long Contexts}

\author{%
  Aditya Sharma\textsuperscript{$*$\hspace{0.1em}{\tiny\faGoogle}} \quad Michael Saxon\textsuperscript{$*$} \quad William Yang Wang \\
  University of California, Santa Barbara\\
  \texttt{\href{https://locovqa.github.io/}{\textbf{LoCoVQA.github.io}}}
}

\begin{document}

\maketitle

\renewcommand{\thefootnote}{\fnsymbol{footnote}}
\footnotetext[1]{Equal contribution. \quad \textsuperscript{\tiny\faGoogle}~Now at Google.}
\renewcommand{\thefootnote}{\arabic{footnote}} %

\begin{strip} 

{
    \vspace{-12.5ex}

    \centerimg[width=\linewidth]{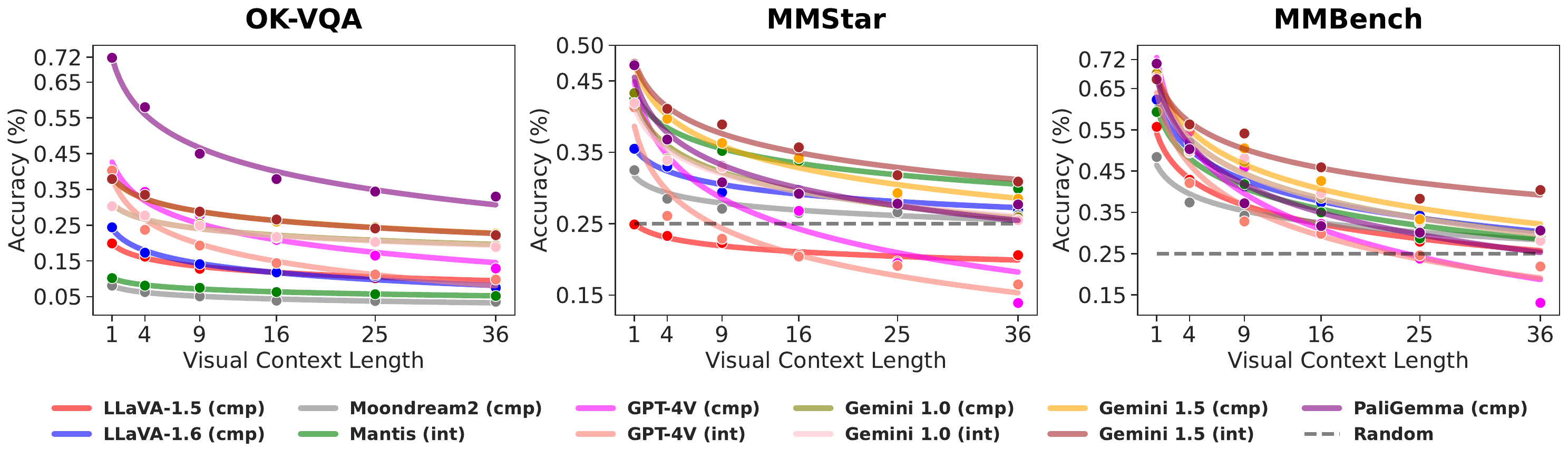}\vspace{-1.5ex}
    \setcounter{figure}{0} \refstepcounter{figure}

    \vspace{-1ex}
      
      \fontsize{10pt}{12pt}\selectfont Figure~\thefigure: The impact of visual context on vision-language models (VLMs) in our modified, multi-image versions of the OK-VQA, MMStar, and MMBench evaluation benchmarks. \textit{Distractor images} around the target image increase the \textit{visual context length} needed to answer the questions. VLM performance exhibits a {logarithmic decay} against distractor count, evident in both single composed (\textbf{cmp}) and multiple interleaved (\textbf{int}) input image configurations.

}
     
\end{strip}

\begin{abstract}
    We present \loco, a dynamic benchmark generator for evaluating long-context extractive reasoning in vision language models (VLMs). \loco~augments test examples for mathematical reasoning, VQA, and character recognition tasks with increasingly long \textit{visual contexts} composed of both in-distribution and out-of-distribution \textit{distractor images}.
    
    Across these tasks, a diverse set of VLMs rapidly lose performance as the visual context length grows, often exhibiting a striking logarithmic decay trend. 
    This test assesses how well VLMs can ignore irrelevant information when answering queries---a task that is quite easy for language models (LMs) in the text domain---demonstrating that current state-of-the-art VLMs lack this essential capability for many long-context applications.

\end{abstract}

\section{Introduction}

Long-context vision-language models (VLMs) which accept multiple image inputs are pushing the boundaries of multimodal reasoning applications, but evaluation methods have not kept up.
As open-weight long-context language models (LMs) have been around for the last few years \cite{dong2023survey}, researchers have developed robust evaluations to compare their capabilities against those of proprietary LMs \cite{reid2024gemini, jiang2024mixtral}.
However, until very recently \cite{jiang2024mantis}, the small set of long-context VLMs supporting multi-image input have been inaccessible to the public \cite{openai2024gpt},
so evaluation practices for long-context VLMs \cite{hong2023cogagent, li2023blip, dai2024instructblip} are in their infancy \cite{zhang2024can}.
Most reasoning-based VLM evals have analogues in text-domain LM tasks.
For example, image captioning \cite{li2022mplug, wang2022git, nguyen2022grit} can be likened to text summarization, and visual question answering (VQA) \cite{chen2022pali, beit, xu2023mplug} to textual question answering (QA).

\captionsetup[subfigure]{labelformat=empty}

\begin{figure*}
    \centering
    \resizebox{0.8\linewidth}{!}{
    \centering
    \begin{tabular}{@{}ccc@{}}
        \fcolorbox{green}{white}{Image ($X$)} & Question ($Q$) & Answer ($A$) \\
        \midrule
         Randomly chosen from dataset $\mathcal{D}$ &
         What cartoon character is on this cake? &
         Winnie the Pooh \\
         \bottomrule
    \end{tabular} } \vspace{1ex}\\
    \centering 
    \resizebox{0.9\linewidth}{!}{
    \centering
    \begin{subfigure}[b]{0.36\textwidth}
        \includegraphics[width=\textwidth]{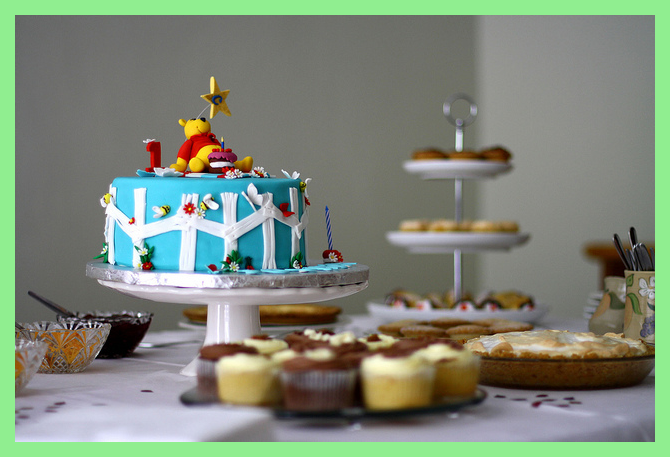}
        \caption{\normalsize 1 image context}
    \end{subfigure}
    \begin{subfigure}[b]{0.28\textwidth}
        \includegraphics[width=\textwidth]{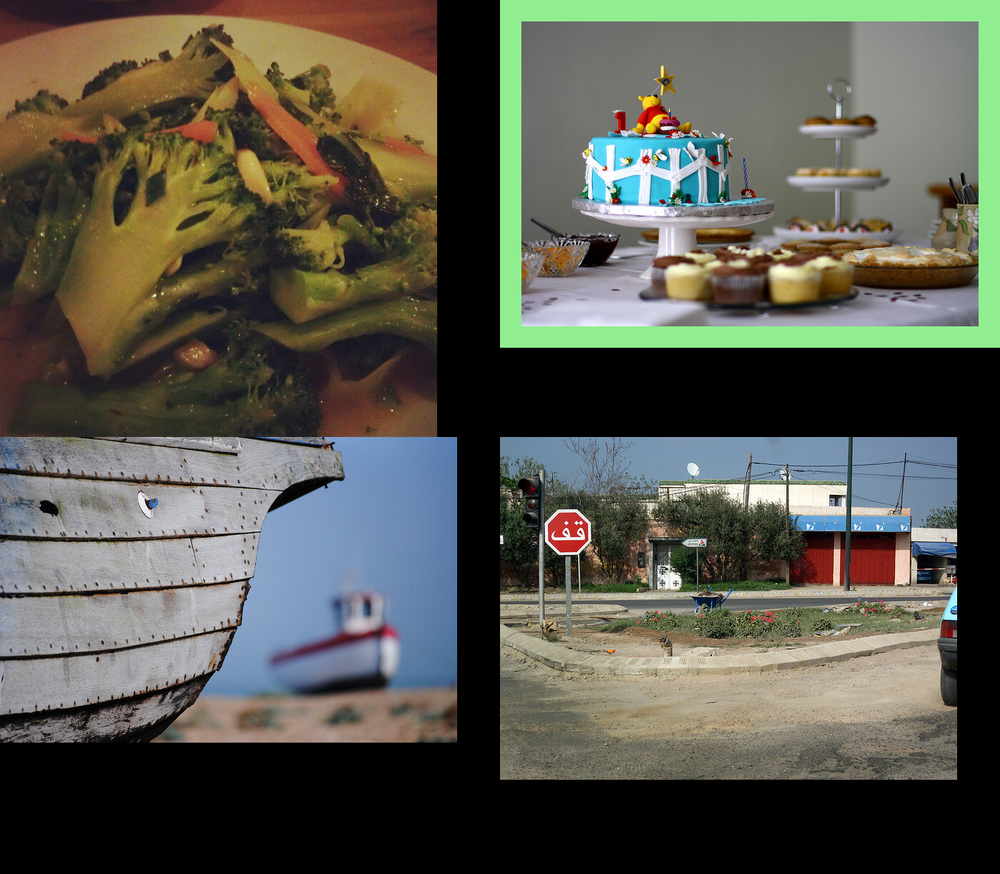}
        \caption{\normalsize 4 image \textit{composed} context}
        \label{fig:4}
    \end{subfigure}
    \begin{subfigure}[b]{0.45\textwidth}
        \includegraphics[width=\textwidth]{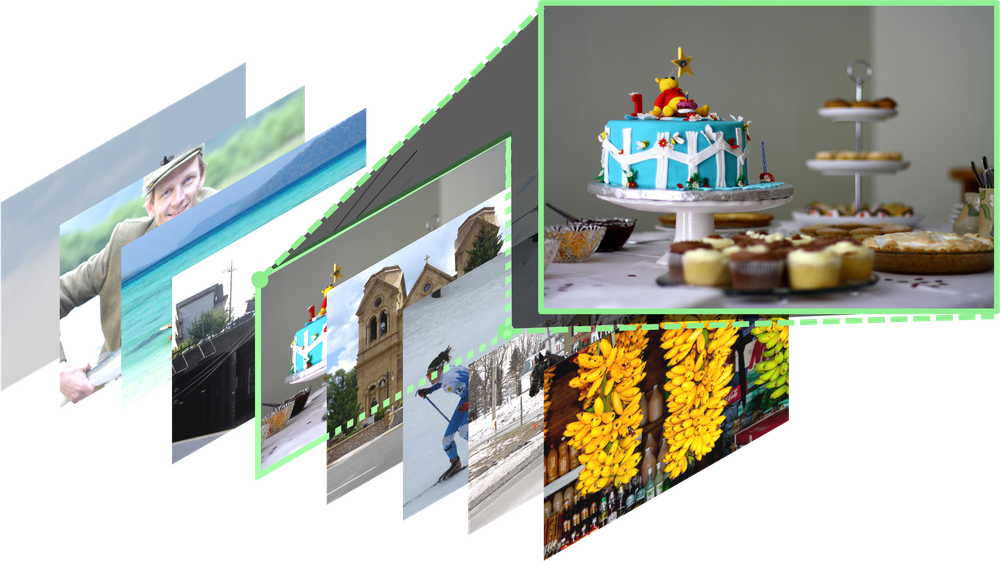}
        \caption{\normalsize 9 image \textit{interleaved} context}
    \end{subfigure} 
   }\vspace{-1ex}
    \caption{Example of \fcolorbox{green}{white}{Image ($X$)} corresponding to question-answer pair ($Q$, $A$) under increasing visual context lengths in the \textit{composed} and \textit{interleaved} settings.   \textit{The green box for illustration only, not included in model inputs.}}
    \label{fig:grid_example}
\end{figure*}

\captionsetup[subfigure]{labelformat=parens}

We aim to bridge gaps in multi-image VLM evaluation by drawing analogies to established long-context LM tasks.
Long-document QA \cite{fan2019eli5} is an easy-to-evaluate test that directly showcases model performance on a useful application, and hints at its potential in retrieval-augmented generation (RAG) more broadly.
Central to these evaluations is the model's ability \textbf{to recognize which elements in a lengthy input are necessary for answering a query and which are not}, a skill we dub \textit{extractive reasoning}. 
Do long-context VLMs also exhibit these extractive reasoning capabilities?

While current VLM benchmarks align with long-document summarization tasks, such as whole-video question answering \cite{huang2021efficient} or summarization \cite{dilawari2019asovs}, no existing benchmarks effectively capture a VLM's ability to \textbf{filter out irrelevant images in a long context to reason or answer a query}---essentially, perform extractive reasoning over images.

Measuring extractive reasoning over image sequences is crucial. Just as extractive textual reasoning facilitates text-domain RAG, multi-modal RAG demands that VLMs efficiently reason over and extract information from documents featuring multiple images.
Similarly, video QA necessitates that models focus solely on frames containing relevant information, much like long-document QA.

We introduce \loco, a benchmark generator for \textbf{Lo}ng \textbf{Co}ntext Visual Question Answering (VQA) \textbf{with distractors}. \loco~enables the creation of long-context image understanding evaluations from any image comprehension dataset by presenting a sequence with the question-relevant image alongside a configurable set of visual distractors.
This allows us to accurately assess how effectively VLMs extract \textbf{only the pertinent information to a query within a cluttered context}. 

We find that even top proprietary VLMs struggle with this capability, even over short visual contexts, likely due to fundamental deficiencies in their training objectives.
By unveiling this \loco~identifies a crucial area for performance improvement in future VLMs.
In summary, we:
\begin{itemize}[itemsep=-5pt, leftmargin=*]
\item Evaluate long-context extractive reasoning across a wide array of VLMs using \loco.
\item Find that all existing VLMs exhibit significant fragility with increasing distractor context sizes.
\item Use \loco~to create a \textit{visual needle in a haystack} test, revealing substantial positional biases in SOTA VLMs in extractive reasoning. 
\end{itemize}

\section{Related Work}

Text-based long-context tasks such as long-document question-answering (QA) \cite{Chen2023UnderstandingRA}, summarization \cite{Phang2022InvestigatingEE}, and retrieval-augmented generation (RAG) \cite{Lewis2020RetrievalAugmentedGF,Xu2023RetrievalML} offer analogues to long-context reasoning in VLMs. %
Many of these tasks (e.g., QA, RAG) require natural language extractive reasoning, as discussed above.

Few VLM long-context evals measure extractive reasoning.
Existing VQA benchmarks involving \textit{distractor information} (thereby extraction) do not focus on long contexts, and long-context VQA benchmarks do not involve distractors. MultipanelVQA \cite{fan2024muffin} includes distractor information, but only within single input images. 
Some VLM evaluations focused on hallucinations indirectly capture a notion of distraction \cite{ghosh2024vdgd, favero2024multi} but do not explicitly measure it alone. 
MILEBench \cite{song2024milebench} evaluates VQA in long contexts, but only on non-distractor tasks such as video summarization or difference detection, where all inputs are relevant.

\textit{Needle-in-a-haystack} evaluation tasks (asking models to recover hidden passphrases at various positions) do require long-context extraction, but reasoning is not involved.
Gemini 1.5 \cite{reid2024gemini} extended this task to video comprehension by hiding a simple passphrase in many positions within a video.
\citet{wang2024needle} present a static benchmark concurrently to ours centered on multimodal needle-in-a-haystack.

\begin{figure*}
    \centering
    \resizebox{0.9\linewidth}{!}{
    \begin{subfigure}[b]{0.27\textwidth}
        \includegraphics[width=\textwidth]{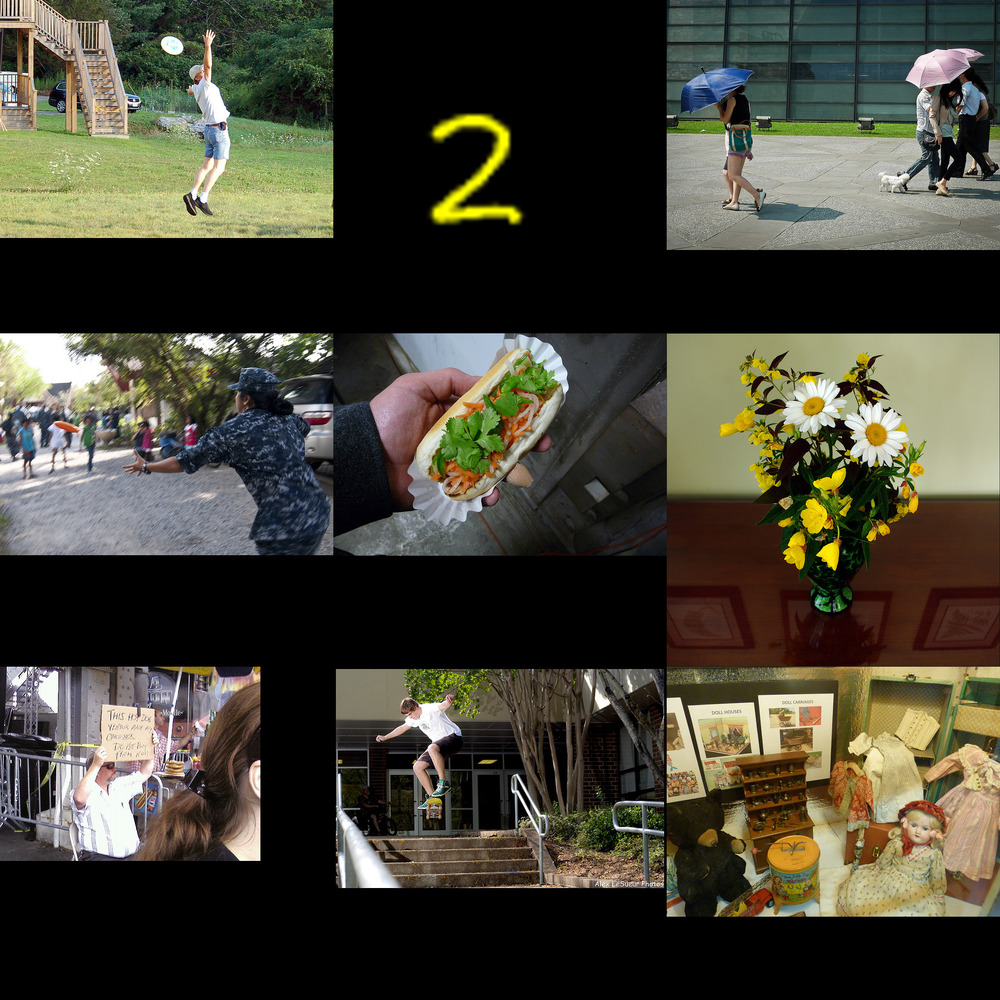}
        \caption{1 digit: \texttt{[2]}}
        \label{fig:mnist1}
    \end{subfigure}
    \begin{subfigure}[b]{0.359\textwidth}
        \includegraphics[width=\textwidth]{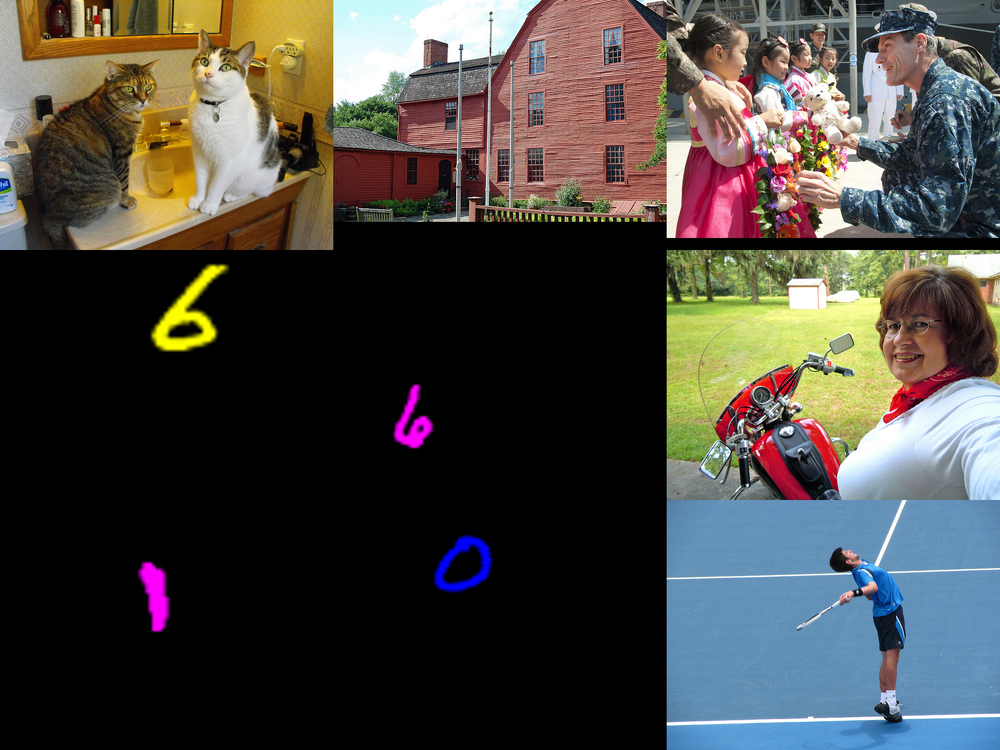}
        \caption{4 digits: \texttt{[6, 6, 1, 0]}}
        \label{fig:mnist4}
    \end{subfigure}
    \begin{subfigure}[b]{0.358\textwidth}
        \includegraphics[width=\textwidth]{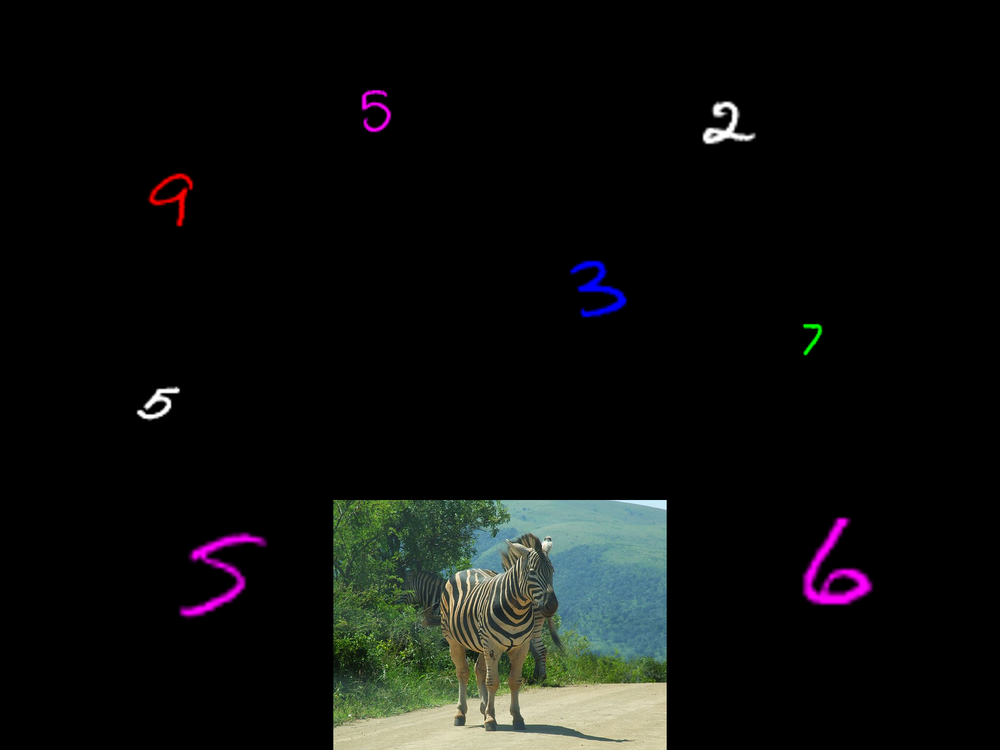}
        \caption{8 digits: \texttt{[9, 5, 2, 5, 3, 7, 5, 6]}}
        \label{fig:mnist9}
    \end{subfigure}
    }
    \caption{Each subfigure represents a variable number of hidden MNIST digits in a 9 image composed context.}
    \label{fig:mnist_example}
\end{figure*}

    \section{\loco~ Generation Method}

Samples synthesized through \loco~ contain one or more \textit{content} images $X$ corresponding to a question and answer pair ($Q$, $A$). Content images can be sampled from various image comprehension benchmark, such as OK-VQA \cite{okvqa}, MMStar \cite{MMStar}, MNIST \cite{lecun1998gradient}, and others. Alongside the content image(s), each sample includes up to 35 \textit{distractor images}, which are either sampled in-distribution from the content image set (ensuring no content image collisions to prevent ambiguity, as described in \S \ref{subsec:filter}) or out-of-distribution from other image sets such as MS COCO \cite{lin2014microsoft}.

Samples can be arranged as \textbf{interleaved} image sequences for VLMs that accept multi-image inputs or as \textbf{composite} images arranged in a grid, as depicted in \autoref{fig:grid_example}. For all models capable of ingesting interleaved sequences, we evaluate both interleaved and composite examples.

\textit{Visual context} refers to the visual elements within an image or sequence of images that are relevant to answering a question. This includes the \textit{content} image and distractor images. 
The challenge is to identify and focus on the pertinent details while ignoring irrelevant information. 
We can sample images both in-distribution to create challenging visual contexts or out-of-distribution to isolate extraction capabilities in the simplest setting. 

As a dynamic generator \loco~ can convert any vision-language dataset into a long-context benchmark \cite{saxon2024benchmarks}. We describe the three datasets we used below.

\subsection{Single-image Reasoning Tasks}\label{subsec:singleimage}

First, we discuss the visual reasoning benchmarks which we expand into long-context samples containing one content image per sample. For most question answering and visual reasoning tasks, this is the only \loco~expansion that makes sense: few VQA samples can be combined such that a plausible new QA pair requiring information from multiple images. Since most interleaved models we evaluate support 36 or fewer images as sequential inputs without modification, we do not evaluate any models with context lengths beyond 36. However, \loco~ scales to any size. For the single-context reasoning tasks, we exclusively sample the distractors in-distribution.

\subsubsection{OK-VQA}

OK-VQA \cite{okvqa} is a single-image visual question answering dataset containing 5,072 question-answer-image triples. It requires external knowledge to reason beyond the image. \loco~generates in-distribution long-context OK-VQA samples, ensuring that no content images have concept collisions that may complicate evaluation. For instance, the question about a character on top of a cake, as shown in \autoref{fig:grid_example}, is sampled from OK-VQA.

Since OK-VQA is an open-domain, free-form answer dataset, we score the samples using three metrics: exact match (full score if the model's response contains any ground truth answer as a substring), and continuous text generation metrics BERTScore \cite{zhang2019bertscore} and ROUGE$_L$ \cite{lin2004rouge} between candidates and references.

\subsubsection{MMStar}

MMStar \cite{MMStar} is a multi-domain visual question answering dataset combining examples from various existing datasets: 424 questions from SEEDBench \cite{seedbench}, 366 questions from MMBench \cite{mmbench}, 100 questions from MMMU \cite{mmmu}, 345 questions from MathVista \cite{mathvista}, 69 examples from ScienceQA \cite{lu2022learn}, and 196 examples from AI2D \cite{kembhavi2016diagram}. MMStar contains 1,500 high-quality multiple-choice questions that \textit{require visual information from the images} to answer, a filtering step not initially performed on the source datasets. For example, over 50\% of ScienceQA questions can be solved by a text-only LLM \cite{MMStar}. Similar to OK-VQA, we generate \loco~samples for MMStar using the collision filtering technique to produce pseudo-documents composed of multiple example images.

As a multiple choice dataset, scoring MMStar is more straightforward. Full details on how we faithfully extract multiple choice answers from the models is provided in \autoref{appendix:scoring}.

\subsubsection{Filtering Collisions in \loco}\label{subsec:filter}

To address the problem of \textit{content-distractor collisions}---where multiple similar in-distribution images in the visual context make the QA pair ambiguous---we implement a robust LM-based filtering method. For each visual context image, we prompt GPT-4 to list the top five entities; if there is overlap, we consider the question potentially ambiguous. Detailed implementations and examples of our filtering method are provided in \autoref{appendix:collision}. To validate this approach, we manually assessed a subset of \loco~generations and found it to be consistent, with no such collisions.

\subsection{Multi-image Reasoning Tasks}\label{subsec:multiimage}

In \S \ref{subsec:singleimage}, we explored tests designed to evaluate whether VLMs can extract a single relevant image from a sequence to answer a query, thereby probing their long-context reasoning capabilities. Extending this to test how well VLMs can extract information from a multi-image sequence is a natural progression. However, VQA examples cannot be easily combined in a way that requires multiple images to answer a single question. Therefore, we turn to constructing ``sequential VQA'' sets using synthetic tasks. Optical character recognition (OCR) is a straightforward task to convert into multi-image question answering by including multiple OCR examples as interleaved images and asking the VLM to list all the text.

\subsubsection{MNIST}

We use MNIST \cite{lecun1998gradient} as it is a canonical dataset for OCR. For a desired visual context length, we sample between 1 and 8 randomly-colored digits from the MNIST training set of around 60K images, resizing them to between 1/6 and 1/2 of the maximum height of other context images. The remaining distractor images are randomly sampled from a subset of 5K high-quality MS COCO \cite{lin2014microsoft} validation images. The VLM is then prompted to \textit{list all handwritten digits present in the sequence}.

By varying the number of digits in the sequence, we can dynamically adjust the difficulty level of the multi-image distractor OCR task. 
\autoref{fig:mnist_example} illustrates examples with 1, 4, and 8 digits in a 9-image context. An output is considered correct only if the stored string of generated digits exactly matches the ground truth, with no partial credit.

\section{Experiments}\label{sec:experiments}

\definecolor{bluehighlight}{RGB}{213, 240, 254}
\definecolor{redhighlight}{RGB}{254, 226, 239}
\definecolor{yellow-mms}{RGB}{255, 253, 84}
\definecolor{red-okvqa}{RGB}{235, 134, 119}
\newcommand{\open}[0]{\rowcolor{bluehighlight}}
\newcommand{\prop}[0]{\rowcolor{redhighlight}}
\newcommand{\coloredsquare}[2][black]{\tikz\fill[#1] (0,0) rectangle (0.65em,0.65em);}

\begin{table}[]
    \centering
    \resizebox{1\linewidth}{!}{
    \begin{tabular}{lcccc}
        \toprule
        \textbf{Vision-Language Model} & \textbf{Context} & \textbf{Base LM} & \textbf{$\downarrow$sample} \\
        \midrule
        \textit{\textbf{Single-Image Input VLMs}} \\
        \open Moondream2-1.6B & 2K & Phi-1.5 & \checkmark \\
        \open LLaVA-1.5-7B & 4K & Vicuna & \checkmark \\
        \open LLaVA-1.6-7B & 8K & Mistral & \checkmark \\
        \open PaliGemma-3B & 8K & Gemma & \checkmark \\
        \textit{\textbf{Multi-Image Input VLMs}} \\
        \open Mantis-Bakllava-7B & 8K & Mistral & \checkmark \\
        \open Mantis-Idefics2-8B & 8K & Mistral & \checkmark \\
        \prop Gemini 1.0 Pro Vision & 32K & Gemini &  - \\ 
        \prop GPT-4 Vision & 128K & GPT-4 &  - \\
        \prop Gemini 1.5 Flash & 1M & Gemini & - \\
        \bottomrule
    \end{tabular}
    }
    \caption{Overview of \coloredsquare[bluehighlight]~ open-source and \coloredsquare[redhighlight]~ proprietary vision-language models (VLMs). \textit{Downsampling} models (\textbf{$\downarrow$sample}) rescale input images in the single-image setting rather than take as many ViT inputs as necessary.
    }
    \label{tab:vlms}
    \vspace{-2ex}
\end{table}

We evaluate the performance of nine current vision-language models on our \loco-generated benchmarks. The open-weight models tested are Moondream2 \cite{moondream2}, LLaVA-1.5 \cite{liu2023improvedllava}, LLaVA-1.6 \cite{liu2024llavanext}, PaliGemma-3b \cite{paligemma}, and two Mantis variants \cite{jiang2024mantis}. Additional details are provided in \autoref{appendix:models}.
Both Mantis variants are VLMs further tuned on interleaved multi-image instruction following. The three proprietary models we evaluate are GPT-4V \cite{achiam2023gpt}, Gemini 1.0 Pro Vision \cite{team2023gemini}, and Gemini 1.5 Flash \cite{reid2024gemini}. \autoref{tab:vlms} showcases all the models along with the model context sizes and image downsampling.

\begin{figure*}[t!]
    \centering
    \vspace{-1ex}
    \includegraphics[width=\textwidth]{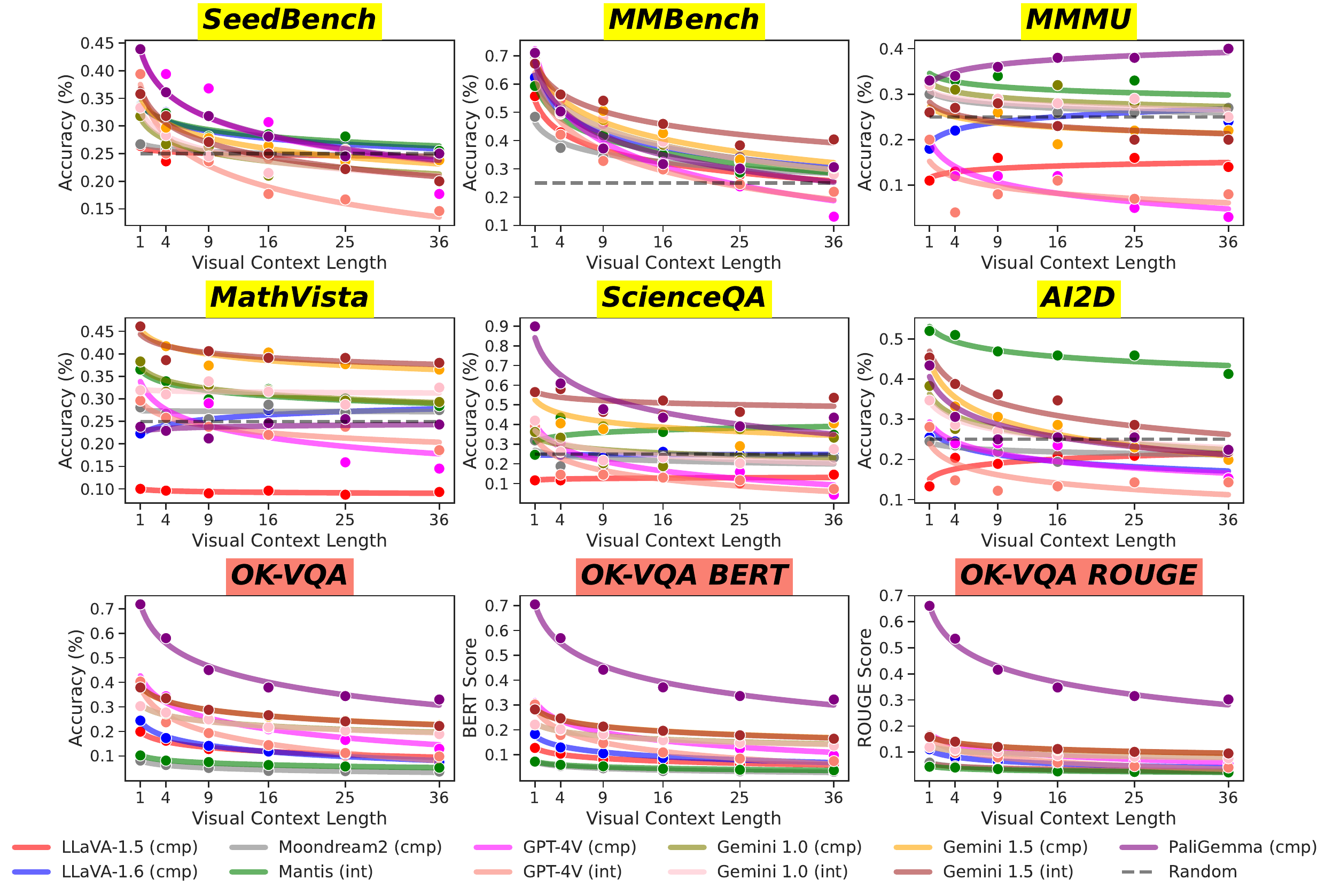}
    \caption{VLM Performance on MMStar and OK-VQA. Note the clearly declining logarithmic fit trends for many of the models. The (model, task) pairs for which these trends do not hold by and large are \textit{below the random baseline.}}
    \label{fig:mmstar-ok-vqa-plot}
    \vspace{-1ex}
\end{figure*}

\paragraph*{Interleaved image support.}
All closed-source models and Mantis support multi-image interleaved inputs, while the others only support single images. For multi-image models, we evaluate both the \textbf{composed (cmp)} \& \textbf{interleaved (int)} settings. For single-image models, we only test the \textbf{composed (cmp)} setting.

\paragraph*{Downsampling images.}
Some models accept images of arbitrary resolution, while others automatically downscale inputs. This difference is crucial, especially in the \textbf{cmp} setting, as increasing the visual context can lead to information loss. For the downsampling models that support both \textbf{cmp} and \textbf{int} settings, we assess both. Any performance differences between these settings would highlight the impact of downsampling on the \textbf{cmp} setting.

\begin{figure*}
    \centering
    \includegraphics[width=\textwidth]{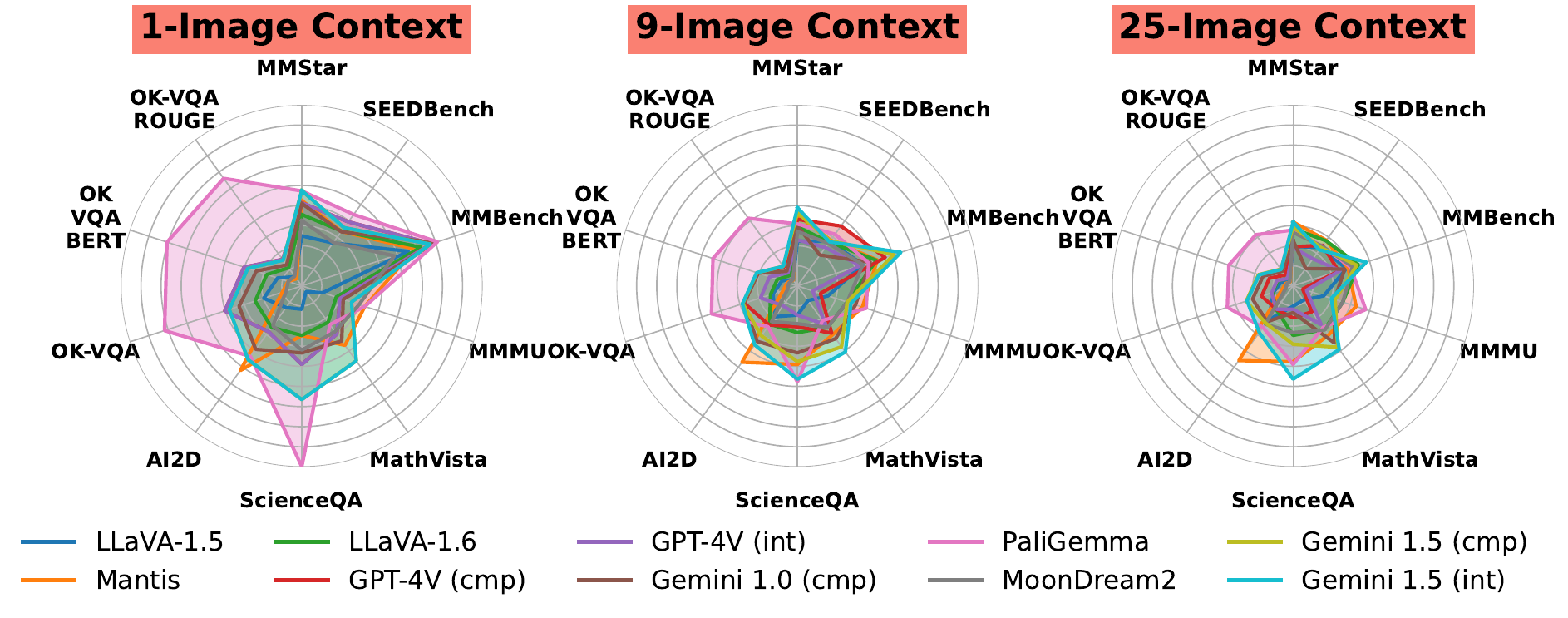}
    \caption{Radar plots of VLM performance across 8 multimodal benchmarks with varied visual context lengths.}
    \label{fig:radar}
\end{figure*}

\section{Results}\label{sec:results}

\newcommand{\hlc}[2][yellow]{{\sethlcolor{#1} \hl{#2}}}
\autoref{fig:mmstar-ok-vqa-plot} illustrates how model performance (across 10 experiments, in both composed and interleaved settings) changes with increasing visual context lengths on single-image \loco~ tasks. The first two rows display results from the MMStar dataset, which contains filtered-image-necessary examples from six other datasets \cite{MMStar}, with titles in\hlc[yellow-mms]{yellow} and random guess thresholds indicated by dotted black lines. The bottom row presents OK-VQA scores using three scoring metrics, with titles in\hlc[red-okvqa]{salmon}.

\begin{figure*}[t!]
    \centering
    \includegraphics[width=\textwidth]{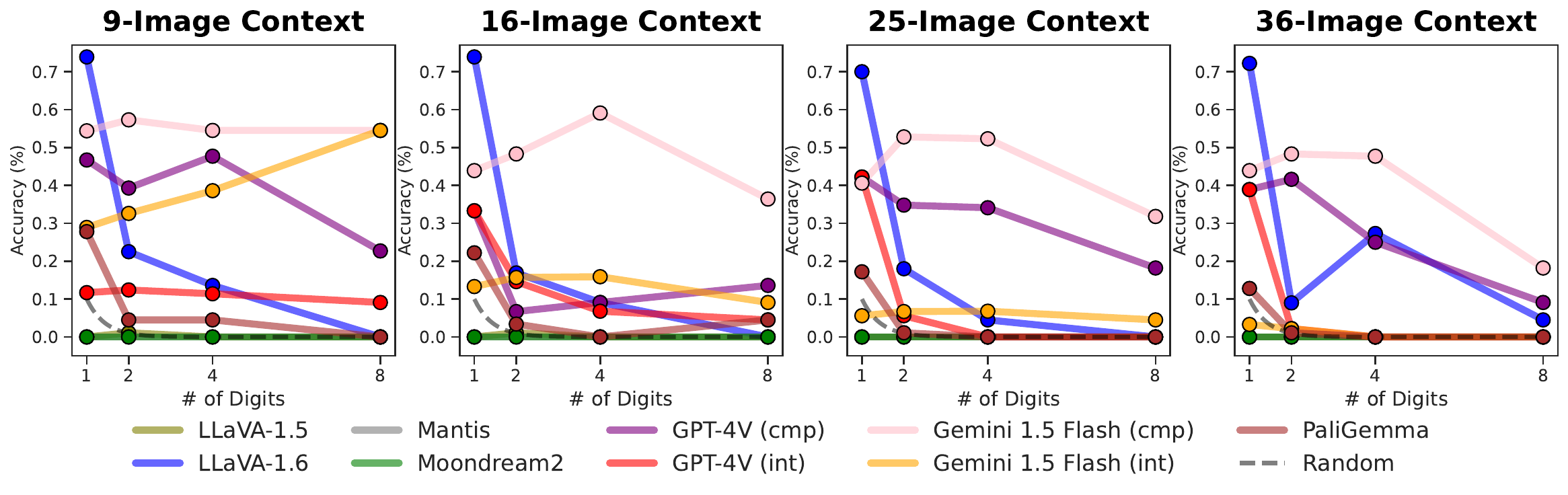}
    \caption{VLM Performance on the MNIST-Digits transcription task as a function of \# of digits to transcribe. \textit{These plots have a different x-axis than the plots in Figure 1 and \autoref{fig:mmstar-ok-vqa-plot}}: rather than the relationship between context size and performance, we are assessing the relationship between "task difficulty" and performance, at four context sizes.}
    \label{fig:mnist-digit-haystack}
    \vspace{-1ex}
\end{figure*}

Across all models on OK-VQA and the majority on MMStar subsets for SeedBench, MMBench, ScienceQA, and AI2D, we observe \textit{striking logarithmic decay trends} in model performance with increasing visual context length. Correlation coefficients, $r^2$, and $p$-values for each trendline are reported in \autoref{tab:rsq-vals}, \autoref{appendix:scoring}.
In general, the closed-source models outperform their open-weight counterparts, especially on multiple-choice tasks. Among the open-weight models, PaliGemma performs the best, likely due to its substantially more extensive training on vision-language tasks compared to its counterparts like LLava or Mantis. 

On several multiple-choice tasks, Mantis (the interleaved open-weight model) outperforms the other open-weight models that rely solely on composite inputs. This is likely due to the image subsampling required by several of the other open-weight models negatively impacting performance when using composite inputs.
However, on the OK-VQA task, Mantis and Moondream are the least performant, even at low context lengths. These models were likely not trained on visual question-answering tasks to the same extent as LLaVA variants during their instruction tuning steps.

The most noteworthy point is this: \textbf{the logarithmic decay trend holds equally well in interleaved and composed settings}. This indicates that the performance-visual context length trend is fundamental and \textbf{cannot be attributed solely to downsampling effects in the composite setting}. Furthermore, for the models tested in both conditions (GPT-4V and Gemini), the same trends are observed in both interleaved and composite settings. Sub-chance performance on the multiple-choice question-answering datasets, particularly for the closed-weight models, occur due to high refusal rates. This may illustrate how ``alignment hampering'' can hinder performance.

\paragraph*{Taskwise Performance by context length.}
\autoref{fig:radar} illustrates the models' performance at context lengths of 1, 9, and 25 for each task, to compare the relative advantages different models have on various tasks. For example, PaliGemma excels on OK-VQA compared to the other models, while Mantis performas well on AI2D. These differences are likely due to variations in training tasks.

\paragraph{Performance on Multi-image Tasks.} 

\autoref{fig:mnist-digit-haystack} presents model performance on the MNIST transcription task by number of digits to transcribe.
\textbf{Note the x-axis in this figure is changed; it is checking for a fixed context length the relationship to number of digits.}
While there is a trend of decreasing performance with an increasing number of images, it does not follow a simple pattern like the context-length trend. For example, Gemini 1.5 experiences minimal performance degradation even as the target digit length increases.

Characterizing difficulty as a function of digit length is complex---as digit counts increase in a fixed context window, \textit{the the output label search space grows}, while the \textit{ratio of relevant images to irrelevant images increases}. 
This may explain why some models have \textit{consistent} or even \textit{increasing} performance as \# digits increases.
Analyzing why different models handle these axes of difficulty differently is an interesting future direction.

However, akin to the single-image tasks, increasing the overall \textit{visual context length} (seen in same-color, x value points across plots), makes the task more difficult, albeit without as clear  a correlation.

\begin{figure*}[t!]
    \centering
    \begin{subfigure}[b]{0.325\textwidth}
        \includegraphics[width=\textwidth]{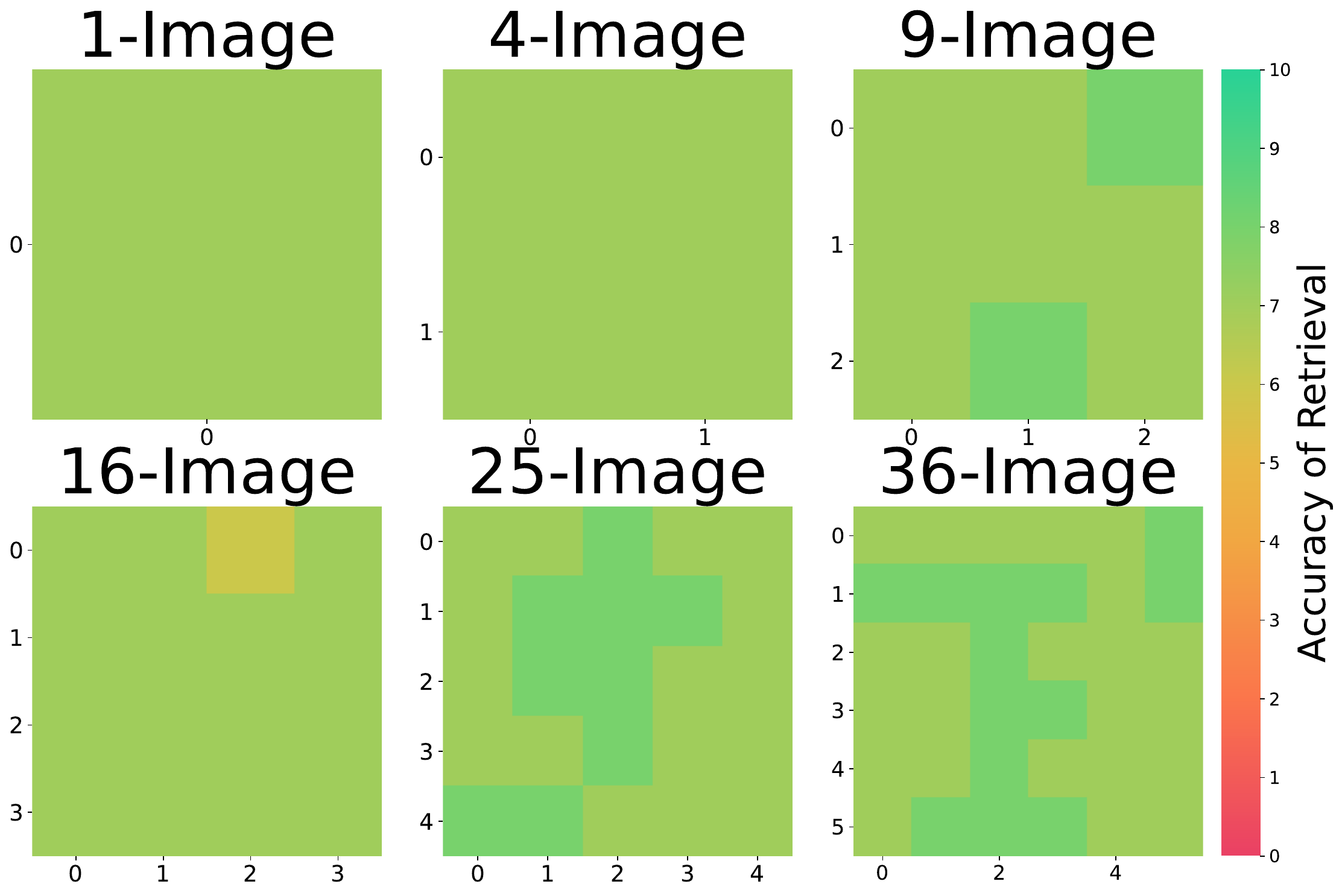}
        \caption{Composed Haystack (LLaVA 1.6)}
        \label{fig:sub1}
    \end{subfigure}
    \begin{subfigure}[b]{0.325\textwidth}
        \includegraphics[width=\textwidth]{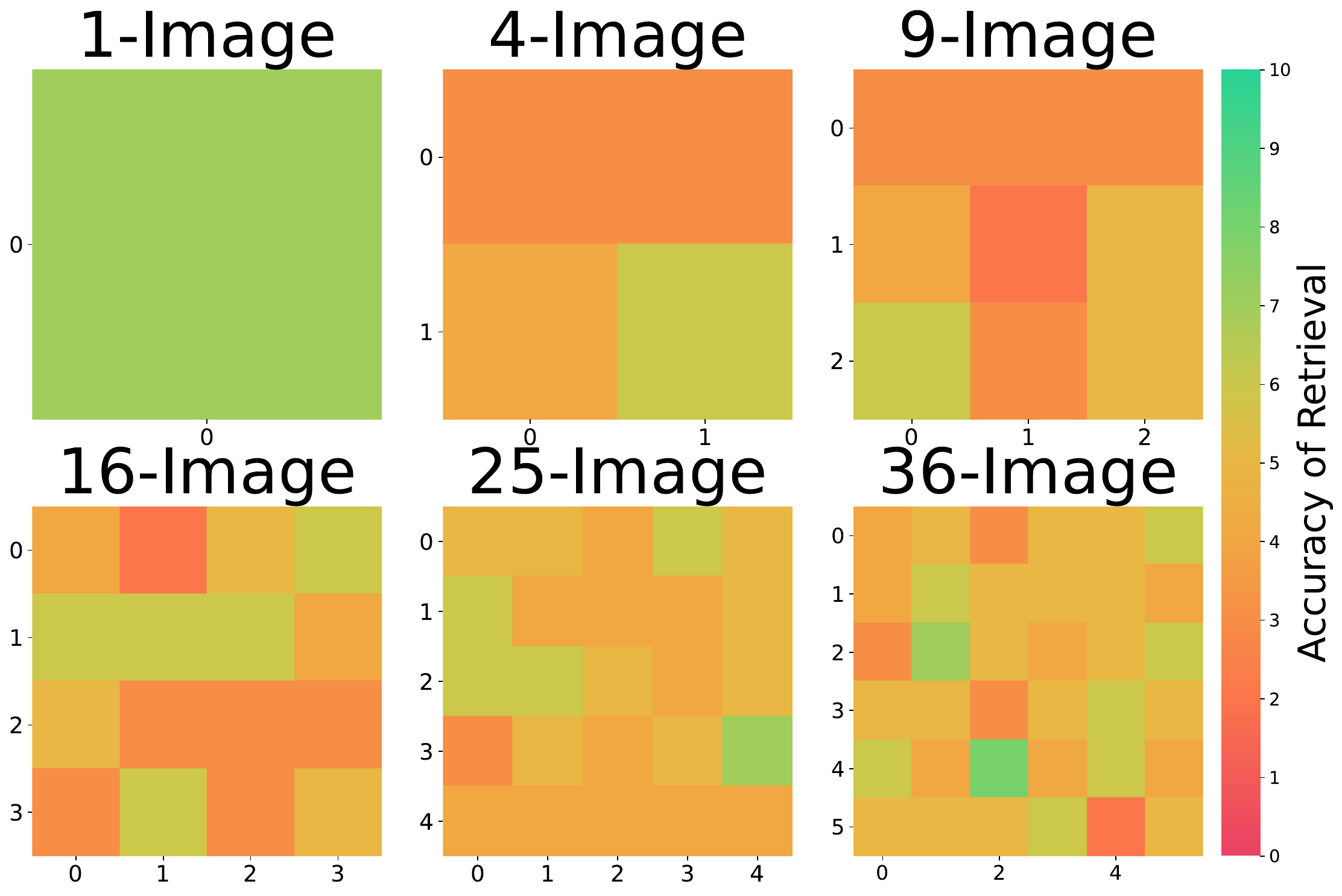}
        \caption{Composed Haystack (GPT-4V)}
        \label{fig:sub4}
    \end{subfigure}
    \begin{subfigure}[b]{0.325\textwidth}
        \includegraphics[width=\textwidth]{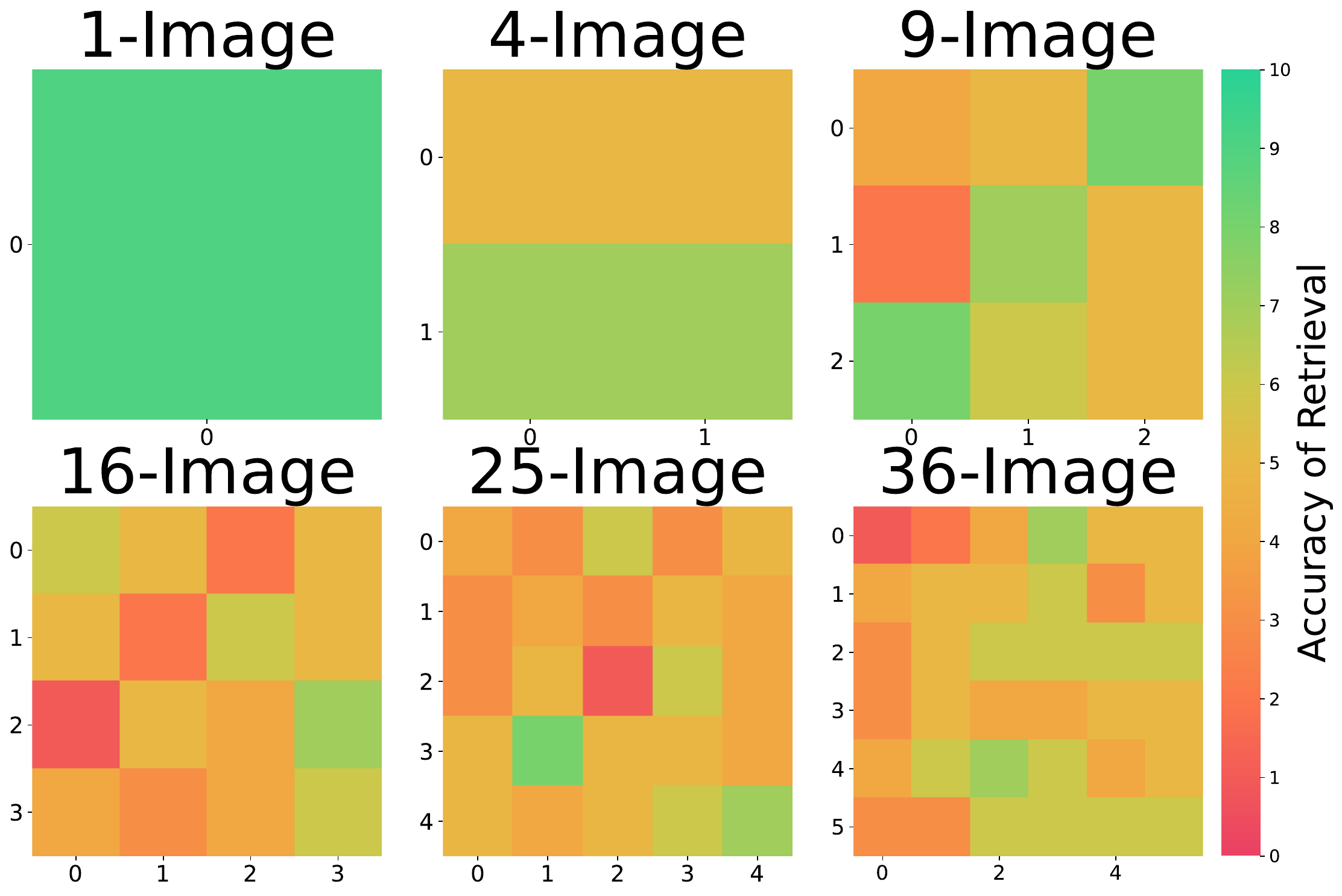}
        \caption{Composed Haystack (Gemini 1.5)}
        \label{fig:sub5}
    \end{subfigure}
    \begin{subfigure}[b]{0.325\textwidth}
        \includegraphics[width=\textwidth]{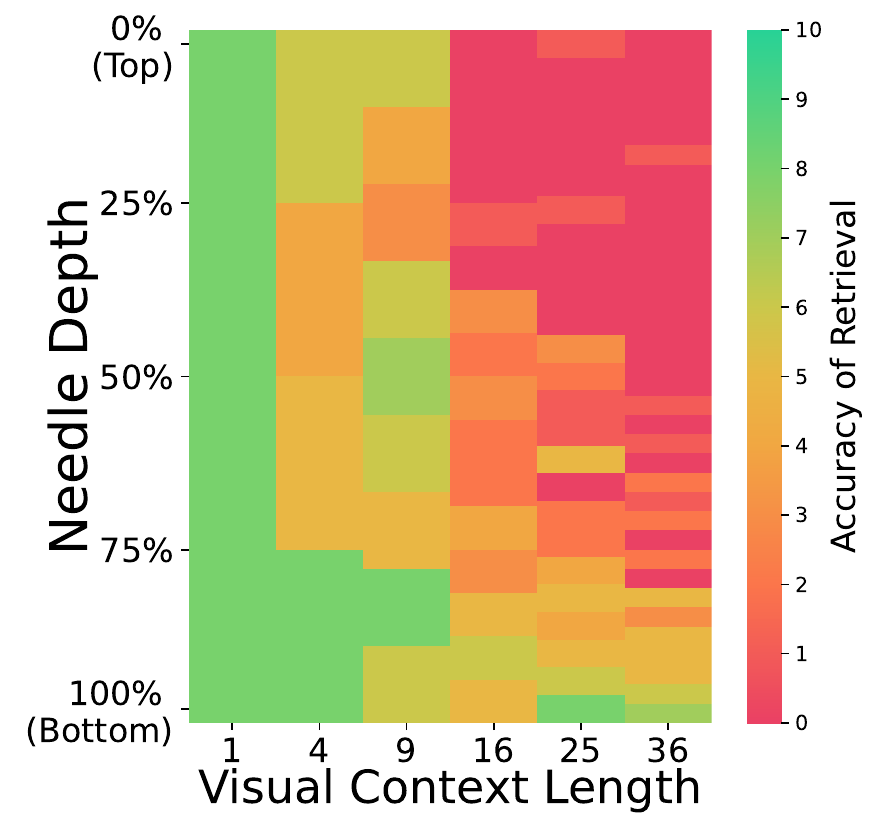}
        \caption{Interleaved Haystack (Mantis)}
        \label{fig:sub2}
    \end{subfigure}
    \begin{subfigure}[b]{0.325\textwidth}
        \includegraphics[width=\textwidth]{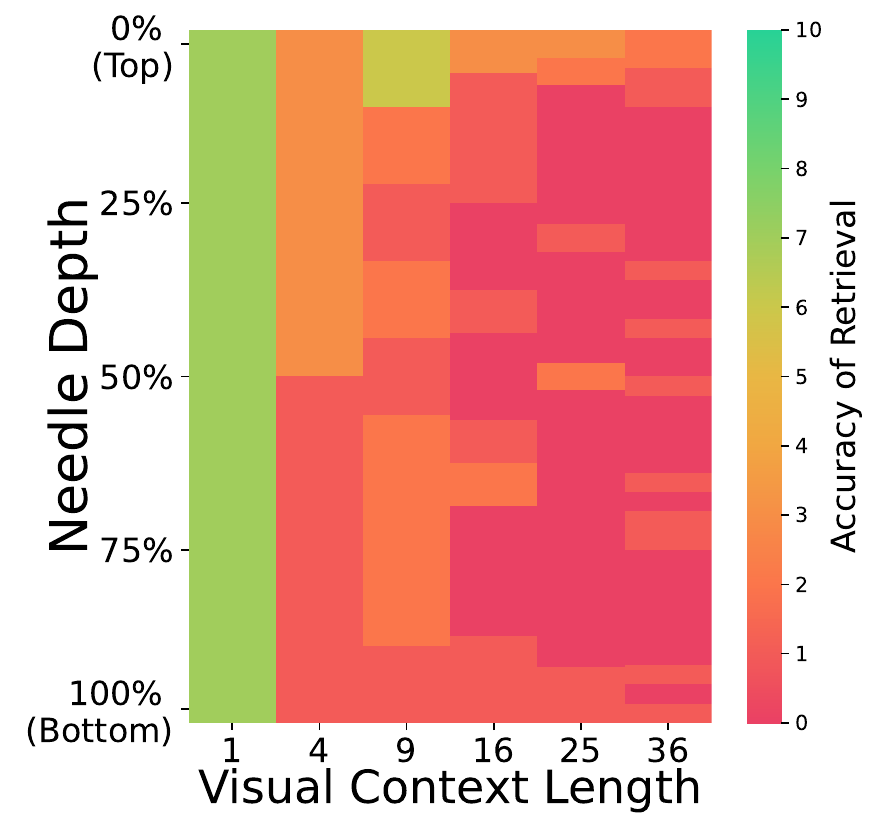}
        \caption{Interleaved Haystack (GPT-4V)}
        \label{fig:sub4}
    \end{subfigure}
    \begin{subfigure}[b]{0.325\textwidth}
        \includegraphics[width=\textwidth]{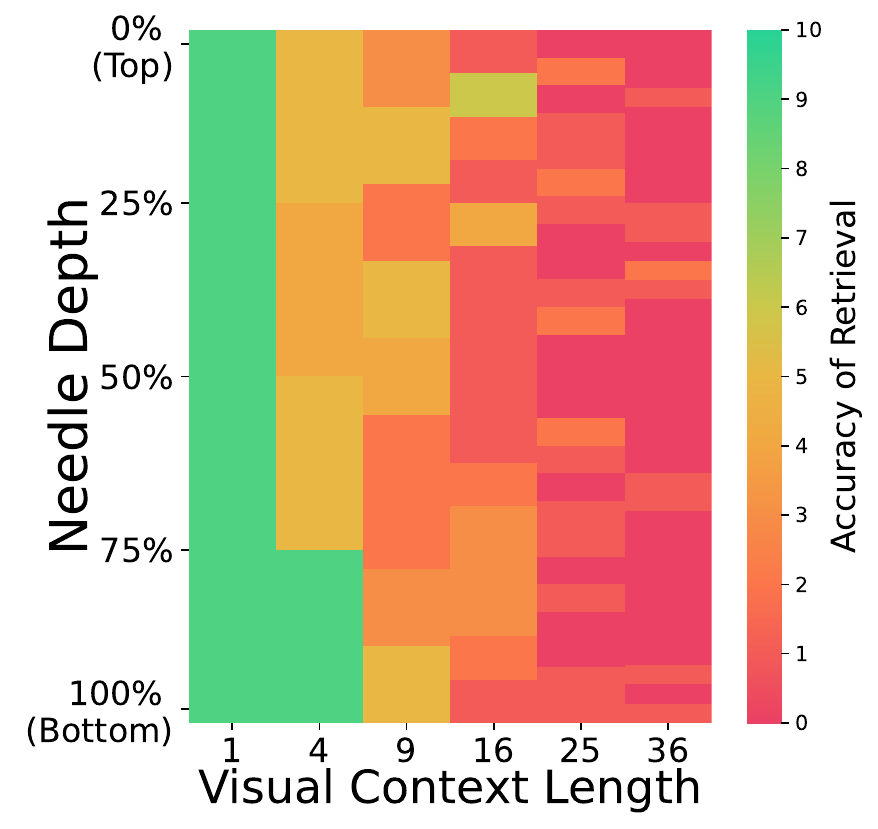}
        \caption{Interleaved Haystack (Gemini 1.5)}
        \label{fig:sub6}
    \end{subfigure}
    \vspace{-1ex}
    \caption{Evaluation of the Visual Needle in a Haystack task using GPT-4V, best-performing VLM, conducted under both composed and interleaved haystack settings. Retrieval accuracy measures the frequency of correct answers produced by the VLM, in this case identifying the MNIST digit. In the composed setting, retrieval accuracy is measured by placing the needle at each cell. In the interleaved setting, needle depth signifies the position within the image sequence, with 0\% representing the first image and 100\% representing the last. Our evaluation highlights a consistent decline in retrieval accuracy as the visual context length increases. %
    }
    \label{fig:needle-in-a-haystack}
    \vspace{-1ex}
\end{figure*}

\section{Ablation: Needle in a Haystack}

As performance within context windows decreases, a natural question arises: \textbf{Are performance failures equally distributed across the visual context range?} To investigate this, we adapt our MNIST-based visual context OCR task into a \textbf{visual needle in a haystack} task.
Needle in a haystack tests are a common minimal test of long-context capabilities in LMs \cite{needleinahaystack}, involving hiding a passphrase, such as the word ``needle,'' in various positions within a long document and asking the model to retrieve it. 
To adapt this concept, we modified our single-digit MNIST task by sampling a set of 10 colored MNIST digits (one for each number, e.g. blue 3, green 7, etc.) and hiding them in each possible position within both interleaved and composed visual context sequences.

By assessing a model's single-digit recovery rate at each position, we can identify any systematic bias in extraction capabilities by postition.
\autoref{fig:needle-in-a-haystack} presents the performance of a subset of the tested VLMs on the composed and interleaved MNIST visual needle in a haystack tasks. We test GPT-4V and Gemini 1.5 Flash in both settings, and treat LLaVA 1.6 and Mantis as analogous methods between the interleaved and composed settings.

Across the three composed tests (subfigures a, b, and c), we do not find a systematic bias toward any particular position on the single-digit recognition task. Surprisingly, LLaVA 1.6 performance the best, probably due to more MNIST-like OCR data in its training mix compared to the others.

However, \textit{we find strong systematic biases with respect to position in the interleaved setting for all models } (subfigures d, e, and f). Mantis shows a preference for late positions in sequences of any length, while GPT-4V and Gemini exhibit weak biases towards early positions. Given that the biases are evident even in the relatively simple MNIST-base OCR task, it is likely that this effect plays a significant role in the performance penalty these models experience with longer contexts.

These results are supported by similar findings in from the image-needle test from \citet{wang2024needle}. 
They also found that---contrary to Google's claims to significant needle-in-a-haystack performance for Gemini \citet{reid2024gemini}---for non-trivial hidden visual information tasks current SOTA VLMs are woefully non-performant.

\section{Discussion}

Why do we observe such striking logarithmic decay in performance with increasing visual context length?
The fact that the trend is consistent in both interleaved and composite settings that downsampling effects in the latter are not the primary cause of performance decay.
There may be an information-theoretic interpretation of this behavior: an increasing signal-to-noise ratio.
As the visual context length increases, the amount of signal (relevant information) remains constant, while the amount of noise (distractor information) increases.

However, this signal-to-noise ratio problem also exists in text-only tasks, yet long-context LLMs perform well on needle in a haystack and long-context QA and summarization tasks, underlying their performance in retrieval-augmented generation. 
When it comes to visual contextual tasks that involve distractor information, VLMs struggle to perform even on the easiest long-context tasks, such as transcribing MNIST characters. They are even less capable on more complex, real-world tasks like distractor VQA.
The core issue appears to be that both short-context and long-context VLMs are not trained on tasks requiring attention over multiple context images to retrieve information. In contrast, for long passages of text, information from throughout the passage remains relevant---referring back to a specific sentence early in a document is intrinsic to the LM training objective.

However, the same may not necessarily be true for the sequential image language modeling task. In the existing interleaved vision-language training corpora, images are likely to followed immediately by their relevant text. As a result, attending to much earlier images in the documents may not be crucial for achieving low LM loss.
Additionally, none of the tested open VLMs are trained on image \textit{generation} tasks conditioned on text while updating the core transformer weights. Overlooking this objective may prevent VLMs from robustly modeling the relationship between images and text, which could be crucial for performing well on these tasks.

\paragraph{Interesting examples of poor performance}

Our analysis unveiled a few particularly surprising results. For example, contrary to our expectation that models capable of handling interleaved input would perform better with it than with composite input, we found that GPT-4V actually favors composite input on many subtasks.
Additionally, to our surprise LLaVA 1.6 performed much better than GPT-4V on the composite haystack task at all sizes.
This outcome was driven by multiple factors: GPT-4V tends to refuses some tasks rather than guessing, while LLaVA always provides a guess. Furthermore, GPT4V often ``hallucinates'' multiple numbers instead of just one.

\paragraph{Possibility of memorization}

Some of the surprising performance results are driven by the training mix. 
For example, Mantis's significant lead on AI2D, and LLaVA's strong performance on the MNIST task may result from having more relevant data or those specific training sets included in their data. 
However, even considering that some models were trained on these tasks, the pronounced drops in generalization performance as the context length increases are even more striking. 
This illustrates that current VLMs fundamentally struggle to attend image sequences as well as they do with text.

\paragraph{Upper bound for video QA}
A more construct-valid test for real-world visual extractive reasoning would be QA over a long video where only a sub-element is required.
E.g., putting the entire film \textit{Star Wars} as input for QA pair (Q: ``which character shoots a green alien in the Mos Eisley cantina?'', A: Han Solo).
Information from a single scene must be extracted to perform this task, thereby requiring extractive reasoning.
However, it is plausible that sequences of related images, unlike sequences of unrelated images, are more in-distribution for long-context VLMs and extractive tasks involving them could be easier for them to solve.
By providing completely unrelated images, our task may be harder than video-based VQA, and may represent an upper-bound for visual extractive reasoning.

\section{Conclusion}

Vision-language models struggle with long-context visual extractive reasoning. Many models fail to extract necessary information from even short contexts of 36, 25, or 16 images, resulting in near- or sub-baseline performance across tasks.
\loco~presents a simple benchmark-generating process applicable to any VQA or retrieval evaluation dataset, making it easy to assess extractive reasoning in VLMs. Our findings suggest that training tasks requiring attention across multiple context images to extract information--rather than simple single-image tasks--should be included in VLM training.
By measuring this capacity it offers an appealing direction for future work.

\section*{Limitations}

Although \loco~is a generalized process applicable to any VLM benchmark, we only evaluated it on three tasks. 
While the strong logarithmic decay trends between visual context length and performance observed across all three are compelling, expansion of \loco~to additional tasks would render the findings clearer.
While \loco~samples distractor images from the same datasets for open-domain and multiple-choice VQA, and our process appears to accurately filter out collisions, a small number of collisions still occur---it is inherently difficult to ensure no failures in an automated generating process \cite{saxon2024lost}. This may lead to a natural ceiling on VLM performance at each context length. 

Other important long-context capabilities likely exist that are not captured by \loco~or prior work such as MILEBench \cite{song2024milebench}. Augmenting these evaluations with tests that capture additional orthogonal VLM long-context capabilities is an important direction for future work.

\section*{Acknowledgements \& Contributions}

Thanks to Yujie Lu, Luke Yoffe, Tejas Srinivasan, Weixi Feng, and Deepak Nathani for discussion and comments.
This work was supported in part by the National Science Foundation Graduate Research Fellowship under Grant No. 1650114, and CAREER Award under Grant No. 2048122.
AS wrote code, ran experiments, and produced figures. 
MS was the primary writer and editor of the manuscript.
Both designed experiments and conducted analysis.

\bibliography{main}

\appendix

\section{Evaluated Models} \label{appendix:models}

We evaluate:
Moondream2-1.6b \cite{moondream2} based on Phi-1.5 \cite{li2023textbooks},
LLaVA-1.5 \cite{liu2023improvedllava} with Vicuna-7b \cite{peng2023instruction} as the LLM backbone,
LLaVA-1.6 (LLaVA-Next) \cite{liu2024llavanext}, an improvement over LLaVA-1.5 with higher image resolution and better visual reasoning, uses Mistral-7b \cite{jiang2023mistral},
PaliGemma-3b \cite{paligemma}, based on open components from the SigLip \cite{zhai2023sigmoid} image encoder and the Gemma \cite{team2024gemma} language model,
Mantis-bakllava-7b \cite{jiang2024mantis} fine-tuned from BakLLaVA \cite{bakllava} and derived from LLaVA but using Mistral-7b \cite{jiang2023mistral},
Mantis-Idefics2-8b \cite{jiang2024mantis}, the current state-of-the-art Mantis variant, based on Idefics2-8b \cite{laurenccon2024matters}.

\section{\loco~ Release Information}

Full source code for generating distract datasets is available at \hyperlink{https://locovqa.github.io}{\texttt{locovqa.github.io}}. \loco~is released under the Apache v2.0 license.

\section{Filtering Collisions in \loco}\label{appendix:collision}
\autoref{fig:appendix-filter} shows an example method used to filter for collisions in 4-Image Context Length input from \autoref{fig:4}. The LLM Query ($Q$) was used to prompt GPT-4 LLM. The second figure shows an example of a collision occurring when creating a 4-image context. In this case, the two images contain oranges, so we resample one of the images to avoid collision. The OK-VQA questions in this case are ``What type of plant do these fruits grow from?'' for the content image and ``In which US states are these fruits commonly grown?'' for the distractor image. Both require visual reasoning step that the fruit described in this image is an orange.

\definecolor{yellowhighlight}{RGB}{253, 242, 220}
\definecolor{greenhighlight}{RGB}{222, 232, 213}
\begin{figure*}
    \begin{subfigure}[b]{0.5\textwidth}
    \centering
    \includegraphics[width=0.99\linewidth]{figures/examples/haystack_4_okvqa.png} 
    \resizebox{\linewidth}{!}{
    \begin{tabular}{ll}
        \toprule
        \multicolumn{2}{c}{LLM Query ($Q$): Please list at most 5 entities in this image} \\
         \midrule
         \rowcolor{yellowhighlight} $X_1$ & Broccoli, Carrot, Pine Nuts, Plate, Garlic \\
         \rowcolor{yellowhighlight} $X_2$ & Boat, Beach, Sky, Sand, Boat Hull \\
         \rowcolor{yellowhighlight} $X_3$ & Cake, Candles, Cupcake, Sandwiches, Dessert Tray \\
         \rowcolor{yellowhighlight} $X_4$ & Traffic Light, Road Sign, Building, Wheelbarrow, Car \\
         \midrule
         \rowcolor{greenhighlight} \ding{51} & Collision Check: $X_1 \cap X_2 \cap X_3 \cap X_4 = \emptyset$ (Valid) \\
         \bottomrule
    \end{tabular} }
    \end{subfigure}%
    ~ 
    \begin{subfigure}[b]{0.5\textwidth}
    \includegraphics[width=0.99\linewidth]{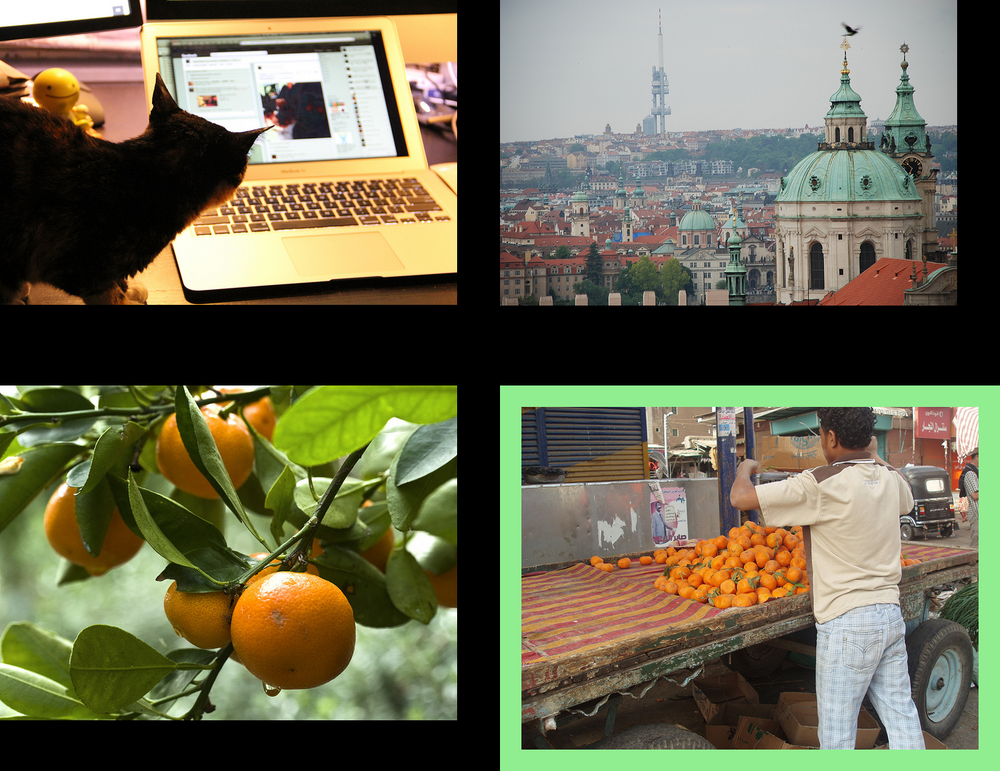}
    \vspace{1ex}\\
    \resizebox{\linewidth}{!}{
    \begin{tabular}{ll}
        \toprule
        \multicolumn{2}{c}{LLM Query ($Q$): Please list at most 5 entities in this image} \\
         \midrule
         \rowcolor{yellowhighlight} $X_1$ & Cat, Laptop, Yellow Figurine, Keyboard, Screen \\
         \rowcolor{yellowhighlight} $X_2$ & Oranges, Leaves, Branches, Water Droplets \\
         \rowcolor{yellowhighlight} $X_3$ & Bird, Dome, Buildings, Television Tower, Trees \\
         \rowcolor{yellowhighlight} $X_4$ & Man, Oranges, Flatbed Cart, Street, Cardboard \\
         \midrule
         \rowcolor{redhighlight}  \ding{55} & Collision Check: $X_1 \cap X_2 \cap X_3 \cap X_4 = \{$Oranges$\}$ \\
         \bottomrule
    \end{tabular}}
    \end{subfigure}
    \caption{Collision Filtering Method for \loco~ by prompting LLM with query $Q$ to identify entities. Cell with \coloredsquare[yellowhighlight]~ represents the entities for each image $X_i$. If there are no entities in common, there are no collisions so we mark cell with \coloredsquare[greenhighlight]~ indicating this is a valid construction of images for \loco.}
    \label{fig:appendix-filter}
\end{figure*}

\section{Model-level Scoring Details}\label{appendix:scoring}

\paragraph{OK-VQA.}
For the OK-VQA free-form ground truth answers, if any of the ground truth candidate answers is a substring of the model-generated answer, we award full points for the exact-matching setting. The other two metrics we used were BERTScore and ROUGE scores. BERTScore \cite{zhang2019bertscore} is a robust text comparison metric which matches candidate and reference based on the cosine similarities of the embeddings. Using the Sentence Transformer package \texttt{all-MiniLM-L6-v2} model, we calculated the maximum BERTScore between all the ground truth answers against the model answer. For the ROUGE-score, we compared the ground truth answer and model answer using the default settings with ROUGE-L, which measure the longest common subsequence at the sentence-level

\paragraph{MMStar.}
For multiple-choice answer form in MMStar, when prompting the VLM, we ask it to produce the answer in the following format: ``Please provide the answer in format \texttt{<answer> </answer>} where the answer is between the tags. For example if the result is apple, your answer should be \texttt{<answer>}apple\texttt{</answer>}.'' This format ensure some grammatical structure and requires the answer to be enclosed within the \texttt{<answer>} tags. GPT4V, Gemini 1.0, and Gemini 1.5 under both composed and interleaved settings produced answers following this format. However, we did not observe consistent behavior from LLaVA-based and Mantis-based variants. Moondream and Gemma consistently responded with a single-choice answer, making the evaluation of these two models the easiest. Due to the variance in VLM responses, we adopted a robust evaluation procedure. First, we checked if the answer was between the \texttt{<answer>} tags and, if so, extracted the MCQ choice directly from the tag. If not, we noted that outputs often followed the format, ``Answer: \textit{choice},'' where the choice follows directly after. We also checked edge cases, including instances where the first letter in the string is a multiple-choice followed by a colon, choices provided in parentheses, only the answer text without the corresponding letter, and a single letter provided.

\paragraph{MNIST.}
For MNIST evaluation, we ensure that the model response contains a list of digits separated by commas. If the output is separated by spaces, we parse it into an array to provide a completely fair evaluation. The response must contain exactly the same number of digits as those in the MNIST digits. We sort both the candidate and reference lists and compare them to check for equality.

\begin{figure*}
    \vspace{-2ex}
    \centering
    \includegraphics[width=0.6\textwidth]{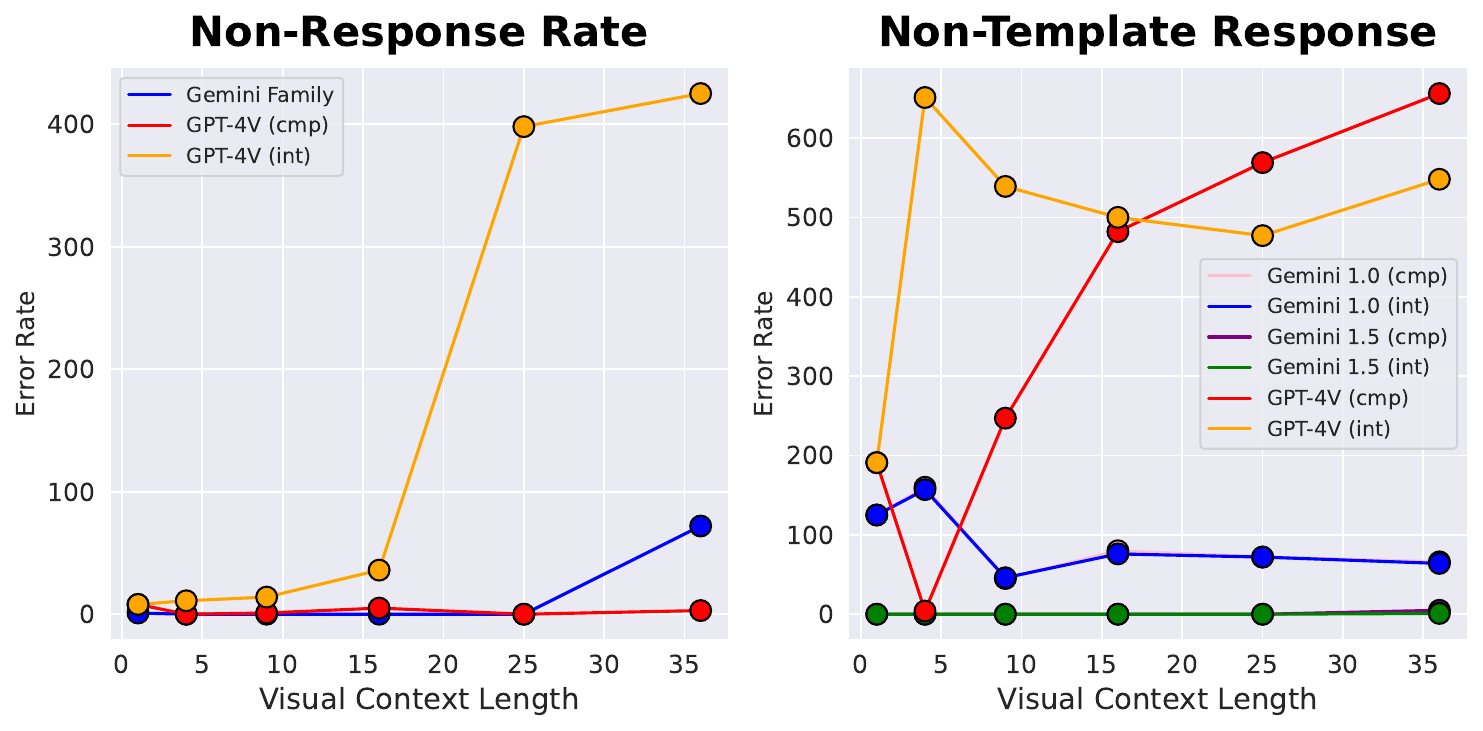}
    \caption{Plots of non-response rate and non-template responses of closed-source VLMs.}
    \label{fig:err-rate}
\end{figure*}

\begin{figure*}
    \centering
    \includegraphics[width=\textwidth]{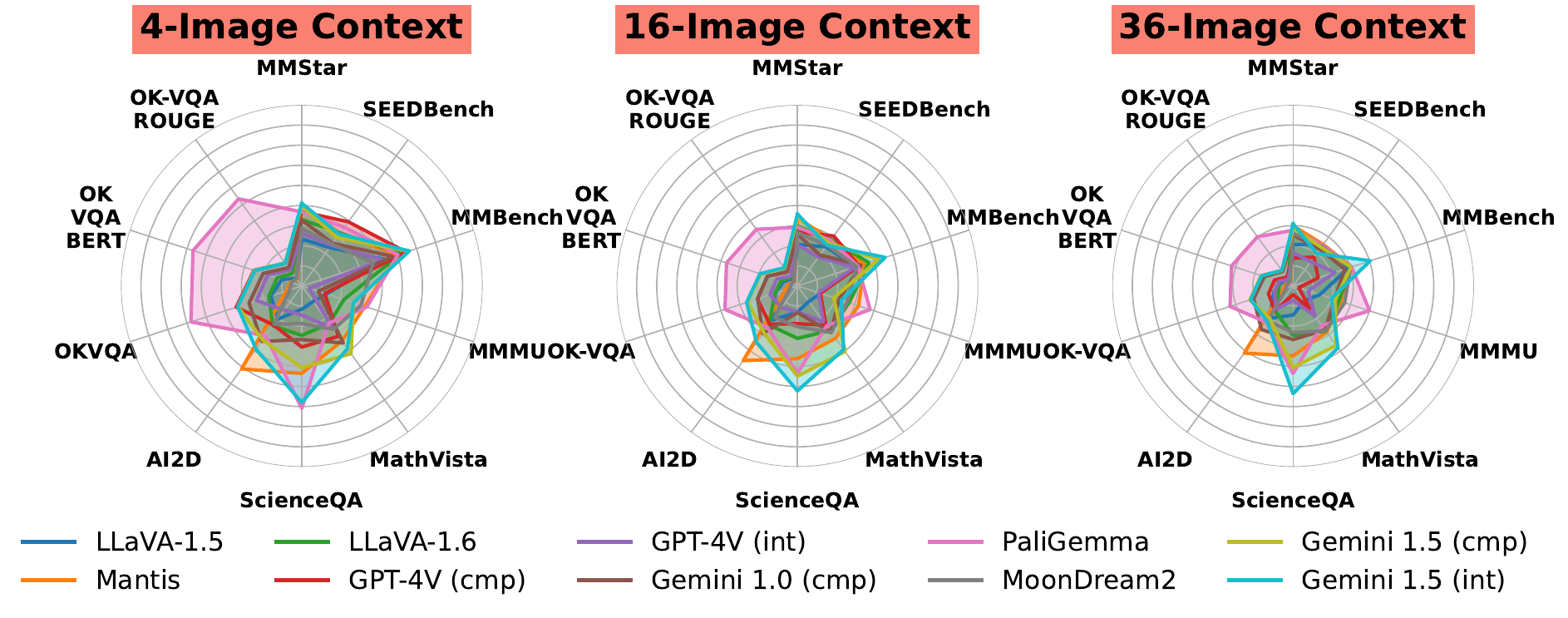}
    \vspace{-4ex}
    \caption{Radar plots of VLM performance across 8 multimodal benchmarks with context lengths $k=4,16,36$.}
    \label{fig:radar-apdx}
    \vspace{-1ex}
\end{figure*}

\paragraph{API model considerations.}
Open-weight models consistently attempt to provide an answer when prompted, even if it means occasionally producing halluncinated responses. In contrast, API-based models like Gemini and GPT-4V sometimes refuse to answer questions under certain circumstances--such as low confidence, potentially unsafe language in the response, or failure to adhere to the response template. We categorize these instances into two primary error types: non-response rate and non-template response rate. \autoref{fig:err-rate} illustrates these error types for both Gemini and GPT-4V models on OK-VQA dataset of size 5072. 

A non-response case is identified when the model returns an empty string. The Gemini models (1.0 and 1.5) exhibit the highest non-responses rates, recording 72 non-responses in visual context with a size of 36, but fewer than 10 for all other context lengths. This pattern may be attributed to the safety mechanisms employed by Gemini or the model's uncertainity due to the presence of multiple distractors. Meanwhile, the non-response rate for GPT-4V in the interleaved setting shows a distinct increasing trend, possibly because GPT-4V struggles with processing a set of 36 interleaved images. The non-response rate for GPT-4V interleaved is 425 non-response in visual context of 36.

A non-template response occurs when the model fails to enclose its output with \texttt{<answer>} tags. This is not penalized in our evaluations--where we still calculate exact match, BERT, ROUGE scores--it is crucial to monitor the trend. As visual context length increases VLMs often struggle to adhere to the prompted response formatting. 
Among the models tested, Gemini 1.5 demonstrates the best adherence to requested formatting, with only a single error. Gemini 1.0 exhibits slightly more issues, with no more than 100 errors in template adherence. 
GPT-4V shows the most significant struggle, with up to 600 formatting errors. 
We classify model refusals, possibly due to low confidence, as failures.

\section{Supplementary Results} \label{appendix:results}
\autoref{fig:radar-apdx} extends \autoref{fig:radar} for context lengths $k=4,16,36$. %
\autoref{tab:rsq-vals-open} and \autoref{tab:rsq-vals} display the $r^2$ values for \autoref{fig:mmstar-ok-vqa-plot} fits, with $p$-val asterisks denoting statistical significance. Red highlighting signifies subchance performance (more than half of scores below random choice). 
Down arrows indicate positive correlation (improved performance with increasing context counters overall trend).

\definecolor{verylightred}{RGB}{255,245,245}
\newcommand{\negcorr}[0]{$^\downarrow$}          %
\newcommand{\subchance}[0]{\cellcolor{verylightred}~} %
\newcommand{\three}{$^{***}$~}
\newcommand{\two}{$^{**}$~}
\newcommand{\one}{$^*$~}

\pgfkeys{
    /LLaVA/.cd,
    0/.initial  = .924\three\subchance,
    1/.initial  = .119\subchance,
    2/.initial  = .959\three,
    3/.initial  = .367\negcorr\two\subchance,
    4/.initial  = .513\one\subchance,
    5/.initial  = .080\negcorr\subchance,
    6/.initial  = .658\negcorr\two\subchance,
    7/.initial  = .985\three,
    8/.initial  = .986\three,
    9/.initial  = .988\three,
    10/.initial = .067\subchance,
    11/.initial = .000\subchance,
    12/.initial = .000\subchance,
    13/.initial = .000\subchance,
}

\pgfkeys{
    /Mantis/.cd,
    0/.initial  = .967\three,
    1/.initial  = .885\three,
    2/.initial  = .972\three,
    3/.initial  = .282\three,
    4/.initial  = .848\three,
    5/.initial  = .209\negcorr\one,
    6/.initial  = .857\three,
    7/.initial  = .991\three,
    8/.initial  = .986\three,
    9/.initial  = .918\three,
    10/.initial = .000\subchance,
    11/.initial = .000\subchance,
    12/.initial = .000\subchance,
    13/.initial = .000\subchance,
}

\pgfkeys{
    /LLaVA15/.cd,
    0/.initial  = .957\three,
    1/.initial  = .825\three,
    2/.initial  = .992\three,
    3/.initial  = .691\negcorr\three\subchance,
    4/.initial  = .907\negcorr\three\subchance,
    5/.initial  = .020\subchance,
    6/.initial  = .604\two,
    7/.initial  = .989\three,
    8/.initial  = .973\three,
    9/.initial  = .973\three,
    10/.initial = .850,
    11/.initial = .790,
    12/.initial = .804\one,
    13/.initial = .596,
}

\pgfkeys{
    /GPT4Vc/.cd,
    0/.initial  = .880\three,
    1/.initial  = .737\three,
    2/.initial  = .951 \three,
    3/.initial  = .861\three\subchance,
    4/.initial  = .676\three\subchance,
    5/.initial  = .932\three\subchance,
    6/.initial  = .743\three\subchance,
    7/.initial  = .963\three,
    8/.initial  = .969\three,
    9/.initial  = .958\three,
    10/.initial = .505,
    11/.initial = .366,
    12/.initial = .863\one,
    13/.initial = .841\one,
}

\pgfkeys{
    /GPT4Vi/.cd,
    0/.initial  = .937\three\subchance,
    1/.initial  = .956\three\subchance,
    2/.initial  = .959\three,
    3/.initial  = .372\three\subchance,
    4/.initial  = .858\three\subchance,
    5/.initial  = .820\three\subchance,
    6/.initial  = .635\two\subchance,
    7/.initial  = .975\three,
    8/.initial  = .977\three,
    9/.initial  = .979\three,
    10/.initial = .632,
    11/.initial = .866\one,
    12/.initial = .704,
    13/.initial = .645,
}

\pgfkeys{
    /Geminic/.cd,
    0/.initial  = .972\three,
    1/.initial  = .765\three,
    2/.initial  = .977\three,
    3/.initial  = .493\two,
    4/.initial  = .832\three,
    5/.initial  = .268\one\subchance,
    6/.initial  = .880\three,
    7/.initial  = .962\three,
    8/.initial  = .955\three,
    9/.initial  = .955\three,
    10/.initial = -,
    11/.initial = -,
    12/.initial = -,
    13/.initial = -,
}

\pgfkeys{
    /Geminii/.cd,
    0/.initial  = .975\three,
    1/.initial  = .833\three,
    2/.initial  = .971\three,
    3/.initial  = .392\two,
    4/.initial  = .027,
    5/.initial  = .598\two\subchance,
    6/.initial  = .809\three,
    7/.initial  = .967\three,
    8/.initial  = .963\three,
    9/.initial  = .957\three,
    10/.initial = -,
    11/.initial = -,
    12/.initial = -,
    13/.initial = -,
}

\pgfkeys{
    /Pali/.cd,
    0/.initial  = .947\three,
    1/.initial  = .988\three,
    2/.initial  = .946\three,
    3/.initial  = .923\negcorr,
    4/.initial  = .132\negcorr\subchance,
    5/.initial  = .907\three,
    6/.initial  = .882\three,
    7/.initial  = .968\three,
    8/.initial  = .968\three,
    9/.initial  = .968\three,
    10/.initial = .895\one,
    11/.initial = .535,
    12/.initial = .651,
    13/.initial = .668,
}

\pgfkeys{
    /Moon/.cd,
    0/.initial  = .903\three,
    1/.initial  = .825\three,
    2/.initial  = .937\three,
    3/.initial  = .586\two,
    4/.initial  = .004,
    5/.initial  = .279\one\subchance,
    6/.initial  = .375\subchance,
    7/.initial  = .979\three,
    8/.initial  = .981\three,
    9/.initial  = .985\three,
    10/.initial = .000\subchance,
    11/.initial = .000\subchance,
    12/.initial = .000\subchance,
    13/.initial = .000\subchance,
}

\pgfkeys{
    /Gemini15c/.cd,
    0/.initial  = .985\three,
    1/.initial  = .977\three,
    2/.initial  = .964\three,
    3/.initial  = .487\two\subchance,
    4/.initial  = .840\three,
    5/.initial  = .546\two,
    6/.initial  = .969\three,
    7/.initial  = .991\three,
    8/.initial  = .989\three,
    9/.initial  = .981\three,
    10/.initial = .052,
    11/.initial = .025,
    12/.initial = .118,
    13/.initial = .489,
}

\pgfkeys{
    /Gemini15i/.cd,
    0/.initial  = .981\three,
    1/.initial  = .980\three,
    2/.initial  = .948\three,
    3/.initial  = .533\two\subchance,
    4/.initial  = .690\three,
    5/.initial  = .309\one,
    6/.initial  = .882\three,
    7/.initial  = .989\three,
    8/.initial  = .993\three,
    9/.initial  = .989\three,
    10/.initial = .895\one,
    11/.initial = .256,
    12/.initial = .146,
    13/.initial = .896,
}

\pgfkeys{
    /1stQ/.cd,
    0/.initial = 10.17\%,
    1/.initial = 8.98\%,
    2/.initial = 8.16\%,
    3/.initial = 7.17\%,
    4/.initial = 27.68\%,
    5/.initial = 24.65\%,
    6/.initial = 14.57\%,
    7/.initial = 15.92\%,
    8/.initial = 33.28\%,
    9/.initial = 28.37\%,
    10/.initial = 25.69\%,
    11/.initial = 23.09\%,
}

\pgfkeys{
    /2ndQ/.cd,
    0/.initial = 10.73\%,
    1/.initial = 10.28\%,
    2/.initial = 8.21\%,
    3/.initial = 7.72\%,
    4/.initial = 23.03\%,
    5/.initial = 15.96\%,
    6/.initial = 10.51\%,
    7/.initial = 9.06\%,
    8/.initial = 31.78\%,
    9/.initial = 28.72\%,
    10/.initial = 22.82\%,
    11/.initial = 21.91\%,
}

\pgfkeys{
    /3rdQ/.cd,
    0/.initial = 10.88\%,
    1/.initial = 8.79\%,
    2/.initial = 8.37\%,
    3/.initial = 9.22\%,
    4/.initial = 20.27\%,
    5/.initial = 14.92\%,
    6/.initial = 8.87\%,
    7/.initial = 7.17\%,
    8/.initial = 33.28\%,
    9/.initial = 29.93\%,
    10/.initial = 23.07\%,
    11/.initial = 21.91\%,
}

\pgfkeys{
    /4thQ/.cd,
    0/.initial = 10.96\%,
    1/.initial = 10.04\%,
    2/.initial = 8.64\%,
    3/.initial = 7.75\%,
    4/.initial = 23.90\%,
    5/.initial = 18.92\%,
    6/.initial = 10.12\%,
    7/.initial = 6.96\%,
    8/.initial = 34.07\%,
    9/.initial = 28.60\%,
    10/.initial = 24.36\%,
    11/.initial = 21.81\%,
}

\begin{table*}[t!]
    \centering
    \begin{adjustbox}{center}
    \resizebox{0.7\textwidth}{!}{
    \begin{tabular}{l|ccccccccccc}
    \toprule
    \textbf{Methods} & LLaVA (c) & LLaVA-1.6 (c) & Moondream (c) & PaliGemma(c) & Mantis (i) \\ 
    \midrule
    MMStar      & \pgfkeysvalueof{/LLaVA/0} & \pgfkeysvalueof{/LLaVA15/0} & \pgfkeysvalueof{/Moon/0} & \pgfkeysvalueof{/Pali/0} & \pgfkeysvalueof{/Mantis/0} \\
    ~SEEDBench  & \pgfkeysvalueof{/LLaVA/1} & \pgfkeysvalueof{/LLaVA15/1} & \pgfkeysvalueof{/Moon/1} & \pgfkeysvalueof{/Pali/1} & \pgfkeysvalueof{/Mantis/1} \\
    ~MMBench    & \pgfkeysvalueof{/LLaVA/2} & \pgfkeysvalueof{/LLaVA15/2} & \pgfkeysvalueof{/Moon/2} & \pgfkeysvalueof{/Pali/2} & \pgfkeysvalueof{/Mantis/2} \\
    ~MMMU       & \pgfkeysvalueof{/LLaVA/3} & \pgfkeysvalueof{/LLaVA15/3} & \pgfkeysvalueof{/Moon/3} & \pgfkeysvalueof{/Pali/3} & \pgfkeysvalueof{/Mantis/3} \\
    ~MathVista  & \pgfkeysvalueof{/LLaVA/4} & \pgfkeysvalueof{/LLaVA15/4} & \pgfkeysvalueof{/Moon/4} & \pgfkeysvalueof{/Pali/4} & \pgfkeysvalueof{/Mantis/4} \\
    ~ScienceQA  & \pgfkeysvalueof{/LLaVA/5} & \pgfkeysvalueof{/LLaVA15/5} & \pgfkeysvalueof{/Moon/5} & \pgfkeysvalueof{/Pali/5} & \pgfkeysvalueof{/Mantis/5} \\
    ~AI2D       & \pgfkeysvalueof{/LLaVA/6} & \pgfkeysvalueof{/LLaVA15/6} & \pgfkeysvalueof{/Moon/6} & \pgfkeysvalueof{/Pali/6} & \pgfkeysvalueof{/Mantis/6} \\
    \midrule
    OK-VQA      & \pgfkeysvalueof{/LLaVA/7} & \pgfkeysvalueof{/LLaVA15/7} & \pgfkeysvalueof{/Moon/7} & \pgfkeysvalueof{/Pali/7} & \pgfkeysvalueof{/Mantis/7} \\
    ~~BERT      & \pgfkeysvalueof{/LLaVA/8} & \pgfkeysvalueof{/LLaVA15/8} & \pgfkeysvalueof{/Moon/8} & \pgfkeysvalueof{/Pali/8} & \pgfkeysvalueof{/Mantis/8} \\
    ~~ROUGE     & \pgfkeysvalueof{/LLaVA/9} & \pgfkeysvalueof{/LLaVA15/9} & \pgfkeysvalueof{/Moon/9} & \pgfkeysvalueof{/Pali/9} & \pgfkeysvalueof{/Mantis/9} \\
    \midrule
    Haystack-9  & \pgfkeysvalueof{/LLaVA/10} & \pgfkeysvalueof{/LLaVA15/10} & \pgfkeysvalueof{/Moon/10} & \pgfkeysvalueof{/Pali/10} & \pgfkeysvalueof{/Mantis/10} \\
    Haystack-16 & \pgfkeysvalueof{/LLaVA/11} & \pgfkeysvalueof{/LLaVA15/11} & \pgfkeysvalueof{/Moon/11} & \pgfkeysvalueof{/Pali/11} & \pgfkeysvalueof{/Mantis/11} \\
    Haystack-25 & \pgfkeysvalueof{/LLaVA/12} & \pgfkeysvalueof{/LLaVA15/12} & \pgfkeysvalueof{/Moon/12} & \pgfkeysvalueof{/Pali/12} & \pgfkeysvalueof{/Mantis/12} \\
    Haystack-36 & \pgfkeysvalueof{/LLaVA/13} & \pgfkeysvalueof{/LLaVA15/13} & \pgfkeysvalueof{/Moon/13} & \pgfkeysvalueof{/Pali/13} & \pgfkeysvalueof{/Mantis/13} \\
    \bottomrule
    \end{tabular}
    }
    \end{adjustbox}
    \caption{Logarithmic curve fit $r^2$ values are reported for each open-weight model, followed by a symbol denoting the $p$-value for statistical significance. The symbols represent: \one for $p \leq 0.05$, \two for $p \leq 0.01$, \& \three for $p \leq 0.001$.
    Light pink cells represent correlations for sets of values that fall below chance performance.}
    \label{tab:rsq-vals-open}
\end{table*}

\begin{table*}[t!]
    \centering
    \begin{adjustbox}{center}
    \resizebox{0.83\textwidth}{!}{
    \begin{tabular}{l|ccccccccccc}
    \toprule
    \textbf{Methods} & GPT-4V (c) & GPT-4V (i) & Gemini 1.0 (c) & Gemini 1.0 (i) & Gemini 1.5 (c) & Gemini 1.5 (i) \\
    \midrule
    MMStar      & \pgfkeysvalueof{/GPT4Vc/0} & \pgfkeysvalueof{/GPT4Vi/0} & \pgfkeysvalueof{/Geminic/0} & \pgfkeysvalueof{/Geminic/0} & \pgfkeysvalueof{/Gemini15c/0} & \pgfkeysvalueof{/Gemini15i/0} \\
    ~SEEDBench  & \pgfkeysvalueof{/GPT4Vc/1} & \pgfkeysvalueof{/GPT4Vi/1} & \pgfkeysvalueof{/Geminic/1} & \pgfkeysvalueof{/Geminic/1} & \pgfkeysvalueof{/Gemini15c/1} & \pgfkeysvalueof{/Gemini15i/1} \\
    ~MMBench    & \pgfkeysvalueof{/GPT4Vc/2} & \pgfkeysvalueof{/GPT4Vi/2} & \pgfkeysvalueof{/Geminic/2} & \pgfkeysvalueof{/Geminic/2} & \pgfkeysvalueof{/Gemini15c/2} & \pgfkeysvalueof{/Gemini15i/2} \\
    ~MMMU       & \pgfkeysvalueof{/GPT4Vc/3} & \pgfkeysvalueof{/GPT4Vi/3} & \pgfkeysvalueof{/Geminic/3} & \pgfkeysvalueof{/Geminic/3} & \pgfkeysvalueof{/Gemini15c/3} & \pgfkeysvalueof{/Gemini15i/3} \\
    ~MathVista  & \pgfkeysvalueof{/GPT4Vc/4} & \pgfkeysvalueof{/GPT4Vi/4} & \pgfkeysvalueof{/Geminic/4} & \pgfkeysvalueof{/Geminic/4} & \pgfkeysvalueof{/Gemini15c/4} & \pgfkeysvalueof{/Gemini15i/4} \\
    ~ScienceQA  & \pgfkeysvalueof{/GPT4Vc/5} & \pgfkeysvalueof{/GPT4Vi/5} & \pgfkeysvalueof{/Geminic/5} & \pgfkeysvalueof{/Geminic/5} & \pgfkeysvalueof{/Gemini15c/5} & \pgfkeysvalueof{/Gemini15i/5} \\
    ~AI2D       & \pgfkeysvalueof{/GPT4Vc/6} & \pgfkeysvalueof{/GPT4Vi/6} & \pgfkeysvalueof{/Geminic/6} & \pgfkeysvalueof{/Geminic/6} & \pgfkeysvalueof{/Gemini15c/6} & \pgfkeysvalueof{/Gemini15i/6} \\
    \midrule
    OK-VQA      & \pgfkeysvalueof{/GPT4Vc/7} & \pgfkeysvalueof{/GPT4Vi/7} & \pgfkeysvalueof{/Geminic/7} & \pgfkeysvalueof{/Geminic/7} & \pgfkeysvalueof{/Gemini15c/7} & \pgfkeysvalueof{/Gemini15i/7} \\
    ~~BERT      & \pgfkeysvalueof{/GPT4Vc/8} & \pgfkeysvalueof{/GPT4Vi/8} & \pgfkeysvalueof{/Geminic/8} & \pgfkeysvalueof{/Geminic/8} & \pgfkeysvalueof{/Gemini15c/8} & \pgfkeysvalueof{/Gemini15i/8} \\
    ~~ROUGE     & \pgfkeysvalueof{/GPT4Vc/9} & \pgfkeysvalueof{/GPT4Vi/9} & \pgfkeysvalueof{/Geminic/9} & \pgfkeysvalueof{/Geminic/9} & \pgfkeysvalueof{/Gemini15c/9} & \pgfkeysvalueof{/Gemini15i/9} \\
    \midrule
    Haystack-9  & \pgfkeysvalueof{/GPT4Vc/10} & \pgfkeysvalueof{/GPT4Vi/10} & \pgfkeysvalueof{/Geminic/10} & \pgfkeysvalueof{/Geminic/10} & \pgfkeysvalueof{/Gemini15c/10} & \pgfkeysvalueof{/Gemini15i/10} \\
    Haystack-16 & \pgfkeysvalueof{/GPT4Vc/11} & \pgfkeysvalueof{/GPT4Vi/11} & \pgfkeysvalueof{/Geminic/11} & \pgfkeysvalueof{/Geminic/11} & \pgfkeysvalueof{/Gemini15c/11} & \pgfkeysvalueof{/Gemini15i/11} \\
    Haystack-25 & \pgfkeysvalueof{/GPT4Vc/12} & \pgfkeysvalueof{/GPT4Vi/12} & \pgfkeysvalueof{/Geminic/12} & \pgfkeysvalueof{/Geminic/12} & \pgfkeysvalueof{/Gemini15c/12} & \pgfkeysvalueof{/Gemini15i/12} \\
    Haystack-36 & \pgfkeysvalueof{/GPT4Vc/13} & \pgfkeysvalueof{/GPT4Vi/13} & \pgfkeysvalueof{/Geminic/13} & \pgfkeysvalueof{/Geminic/13} & \pgfkeysvalueof{/Gemini15c/13} & \pgfkeysvalueof{/Gemini15i/13} \\
    \bottomrule
    \end{tabular}
    }
    \end{adjustbox}
    \caption{Logarithmic curve fit $r^2$ values are reported for each closed-source model, followed by a symbol denoting the $p$-value for statistical significance. The symbols represent: \one for $p \leq 0.05$, \two for $p \leq 0.01$, \& \three for $p \leq 0.001$. Light pink cells represent correlations for sets of values that fall below chance performance.}
    \label{tab:rsq-vals}
\end{table*}

\begin{table*}
    \centering
    \begin{adjustbox}{center}
        \resizebox{0.8\textwidth}{!}{
        \begin{tabular}{l|l|cccc}
        \toprule
        \textbf{Models} & \textbf{Image Context} & \textbf{1$^{st}$ Quartile} & \textbf{2$^{nd}$ Quartile} & \textbf{3$^{rd}$ Quartile} & \textbf{4$^{th}$ Quartile} \\
        \midrule
        \textbf{Mantis} & $k=4$ & 
        \pgfkeysvalueof{/1stQ/0} & \pgfkeysvalueof{/2ndQ/0} & \pgfkeysvalueof{/3rdQ/0} & \pgfkeysvalueof{/4thQ/0} \\
        & $k=9$ & 
        \pgfkeysvalueof{/1stQ/1} & \pgfkeysvalueof{/2ndQ/1} & \pgfkeysvalueof{/3rdQ/1} & \pgfkeysvalueof{/4thQ/1} \\
        & $k=25$ &
        \pgfkeysvalueof{/1stQ/2} & \pgfkeysvalueof{/2ndQ/2} & \pgfkeysvalueof{/3rdQ/2} & \pgfkeysvalueof{/4thQ/2} \\
        & $k=36$ & 
        \pgfkeysvalueof{/1stQ/3} & \pgfkeysvalueof{/2ndQ/3} & \pgfkeysvalueof{/3rdQ/3} & \pgfkeysvalueof{/4thQ/3} \\
        \midrule
        \textbf{GPT-4V} & $k=4$ &
        \pgfkeysvalueof{/1stQ/4} & \pgfkeysvalueof{/2ndQ/4} & \pgfkeysvalueof{/3rdQ/4} & \pgfkeysvalueof{/4thQ/4} \\
        & $k=9$ &
        \pgfkeysvalueof{/1stQ/5} & \pgfkeysvalueof{/2ndQ/5} & \pgfkeysvalueof{/3rdQ/5} & \pgfkeysvalueof{/4thQ/5} \\
        & $k=25$ &
        \pgfkeysvalueof{/1stQ/6} & \pgfkeysvalueof{/2ndQ/6} & \pgfkeysvalueof{/3rdQ/6} & \pgfkeysvalueof{/4thQ/6} \\
        & $k=36$ &
        \pgfkeysvalueof{/1stQ/7} & \pgfkeysvalueof{/2ndQ/7} & \pgfkeysvalueof{/3rdQ/7} & \pgfkeysvalueof{/4thQ/7} \\
        \midrule
        \textbf{Gemini 1.5} & $k=4$ &
        \pgfkeysvalueof{/1stQ/8} & \pgfkeysvalueof{/2ndQ/8} & \pgfkeysvalueof{/3rdQ/8} & \pgfkeysvalueof{/4thQ/8} \\
        & $k=9$ &
        \pgfkeysvalueof{/1stQ/9} & \pgfkeysvalueof{/2ndQ/9} & \pgfkeysvalueof{/3rdQ/9} & \pgfkeysvalueof{/4thQ/9} \\
        & $k=25$ &
        \pgfkeysvalueof{/1stQ/10} & \pgfkeysvalueof{/2ndQ/10} & \pgfkeysvalueof{/3rdQ/10} & \pgfkeysvalueof{/4thQ/10} \\
        & $k=36$ &
        \pgfkeysvalueof{/1stQ/11} & \pgfkeysvalueof{/2ndQ/11} & \pgfkeysvalueof{/3rdQ/11} & \pgfkeysvalueof{/4thQ/11} \\
        \bottomrule
        \end{tabular}
        }
    \end{adjustbox}
    \caption{\textit{Impact of ordering on non-haystack tests}---to assess if a positional bias akin to that in the \textit{needle in a haystack} test is present for non-trivial tasks, we separately-averaged the OK-VQA scores for three models by the quartile the content image appears in the sequence. This confirms the same positional biases observed for MNIST haystack tests hold in the more challenging VQA setting.}
\end{table*}

\newpage
\definecolor{questioncolor}{RGB}{255, 230, 204}
\definecolor{answercolor}{RGB}{204, 255, 204}
\definecolor{errorcolor}{RGB}{255, 204, 204}
\definecolor{tableborder}{RGB}{192, 192, 192}
\definecolor{bluehighlight}{RGB}{173, 216, 230}

\cleardoublepage
\onecolumn

\section{Case Study}
\textbf{List of Case Study Figures}
\begin{enumerate}[label=\Roman*]
    \item \textbf{OK-VQA} -- {\hyperref[acs_okvqa]{Example 1}} \\
    Visual Context Length 
    ($k=1$, \hspace{.25em} $k=4$, \hspace{.25em} $k=9$, \hspace{.25em} $k=25$, \hspace{.25em} $k=36$)
    \item \textbf{MMStar (SeedBench)} -- {\hyperref[mmstar_sb]{Example 2}} \\
    Visual Context Length 
    ($k=1$, \hspace{.25em} $k=4$, \hspace{.25em} $k=9$, \hspace{.25em} $k=25$, \hspace{.25em} $k=36$)
    \item \textbf{MMStar (MMBench)} -- {\hyperref[mmstar_mb]{Example 3}} \\
    Visual Context Length 
    ($k=1$, \hspace{.25em} $k=4$, \hspace{.25em} $k=9$, \hspace{.25em} $k=25$, \hspace{.25em} $k=36$)
    \item \textbf{MMStar (MathVista)} -- {\hyperref[mmstar_mv]{Example 4}} \\
    Visual Context Length 
    ($k=1$, \hspace{.25em} $k=4$, \hspace{.25em} $k=9$, \hspace{.25em} $k=25$, \hspace{.25em} $k=36$)
\end{enumerate}

\begin{center}
    \arrayrulecolor{tableborder}
    \begin{tabular}{|>{\centering\arraybackslash}m{0.9\linewidth}|}
        \hline
        \rowcolor[gray]{0.9}\textbf{\large Example 1 -- OK-VQA}
        \addcontentsline{toc}{section}{OK-VQA}
    \end{tabular}
    \begin{tabular}{|>{\raggedright\arraybackslash}m{0.9\linewidth}|}
        \hline
        \rowcolor{questioncolor}\textbf{Question:} What activity is this? \\
        \hline
        \begin{center}
            \fbox{\includegraphics[width=0.4\linewidth]{figures/appedix/okvqa.png}} \\
            \texttt{<image>}
        \end{center} \\
        \hline
        \label{acs_okvqa}
    \end{tabular}
    \begin{tabular}{|>{\centering\arraybackslash}m{0.9\linewidth}|}
        \rowcolor{bluehighlight}\textbf{Visual Context Length ($k=1$)} \label{acs_okvqa_k1}\\
        \begin{tabularx}{0.913\textwidth}{!{\color{tableborder}\vrule} X c !{\color{tableborder}\vrule}}
            \rowcolor{answercolor}\textbf{LLaVA 1.5 (cmp)}: & \textbf{This is a horseback riding activity} \\
            \rowcolor{answercolor}\textbf{LLaVA 1.6 (cmp)}: & \textbf{Horseback riding} \\
            \rowcolor{answercolor}\textbf{Moondream2 (cmp)}: & \textbf{Horseback riding} \\
            \rowcolor{answercolor}\textbf{Mantis (int)}: & \textbf{Horseback riding} \\
            \rowcolor{answercolor}\textbf{GPT-4V (cmp)}: & \textbf{horseback riding} \\
            \rowcolor{answercolor}\textbf{GPT-4V (int)}: & \textbf{horseback riding} \\
            \rowcolor{answercolor}\textbf{Gemini 1.0 (cmp)}: & \textbf{horseback riding} \\
            \rowcolor{answercolor}\textbf{Gemini 1.0 (int)}: & \textbf{horseback riding} \\
            \rowcolor{answercolor}\textbf{Gemini 1.5 (cmp)}: & \textbf{Horseback riding} \\
            \rowcolor{answercolor}\textbf{Gemini 1.5 (int)}: & \textbf{Horseback riding} \\
            \rowcolor{answercolor}\textbf{PaliGemma (cmp)}: & \textbf{horseback ride} \\
        \end{tabularx} \\ \\
        \hline
    \end{tabular}
    \begin{tabular}{|>{\centering\arraybackslash}m{0.9\linewidth}|}
        \rowcolor{bluehighlight}\textbf{Visual Context Length ($k=4$)}\label{acs_okvqa_k4} \\
        \addcontentsline{toc}{subsection}{4-Image Context}
        \begin{tabularx}{0.913\textwidth}{!{\color{tableborder}\vrule} X c !{\color{tableborder}\vrule}}
            \rowcolor{answercolor}\textbf{LLaVA 1.5 (cmp)}: & \textbf{a girl riding a horse} \\
            \rowcolor{errorcolor}\textbf{LLaVA 1.6 (cmp)}: & \textbf{Photography} \\
            \rowcolor{errorcolor}\textbf{Moondream2 (cmp)}: & \textbf{biking} \\
            \rowcolor{errorcolor}\textbf{Mantis (int)}: & \textbf{walking} \\
            \rowcolor{answercolor}\textbf{GPT-4V (cmp)}: & \textbf{horseback riding} \\
            \rowcolor{answercolor}\textbf{GPT-4V (int)}: & \textbf{horseback riding} \\
            \rowcolor{answercolor}\textbf{Gemini 1.0 (cmp)}: & \textbf{Horseback Riding} \\
            \rowcolor{answercolor}\textbf{Gemini 1.0 (int)}: & \textbf{Horseback Riding} \\
            \rowcolor{errorcolor}\textbf{Gemini 1.5 (cmp)}: & \textbf{tennis} \\
            \rowcolor{errorcolor}\textbf{Gemini 1.5 (int)}: & \textbf{tennis} \\
            \rowcolor{answercolor}\textbf{PaliGemma (cmp)}: & \textbf{horseback ride} \\
        \end{tabularx} \\
        \hline
    \end{tabular}
    \begin{tabular}{|>{\centering\arraybackslash}m{0.9\linewidth}|}
        \rowcolor{bluehighlight}\textbf{Visual Context Length ($k=9$)} \label{acs_okvqa_k9} \\
        \addcontentsline{toc}{subsection}{9-Image Context}
        \begin{tabularx}{0.913\textwidth}{!{\color{tableborder}\vrule} X c !{\color{tableborder}\vrule}}
            \rowcolor{answercolor}\textbf{LLaVA 1.5 (cmp)}: & \textbf{woman walking her horse} \\
            \rowcolor{errorcolor}\textbf{LLaVA 1.6 (cmp)}: & \textbf{collage} \\
            \rowcolor{errorcolor}\textbf{Moondream2 (cmp)}: & \textbf{biking} \\
            \rowcolor{errorcolor}\textbf{Mantis (int)}: & \textbf{woman preparing to go surfing} \\
            \rowcolor{errorcolor}\textbf{GPT-4V (cmp)}: & \textbf{bowling} \\
            \rowcolor{answercolor}\textbf{GPT-4V (int)}: & \textbf{horseback riding} \\
            \rowcolor{answercolor}\textbf{Gemini 1.0 (cmp)}: & \textbf{horseback riding} \\
            \rowcolor{answercolor}\textbf{Gemini 1.0 (int)}: & \textbf{horseback riding} \\
            \rowcolor{errorcolor}\textbf{Gemini 1.5 (cmp)}: & \textbf{Bowling} \\
            \rowcolor{errorcolor}\textbf{Gemini 1.5 (int)}: & \textbf{bowling} \\
            \rowcolor{answercolor}\textbf{PaliGemma (cmp)}: & \textbf{horseback riding} \\
        \end{tabularx} \\ \\
        \hline
    \end{tabular}
    \begin{tabular}{|>{\centering\arraybackslash}m{0.9\linewidth}|}
        \rowcolor{bluehighlight}\textbf{Visual Context Length ($k=16$)} \label{acs_okvqa_k16} \\
        \addcontentsline{toc}{subsection}{16-Image Context}
        \begin{tabularx}{0.913\textwidth}{!{\color{tableborder}\vrule} X c !{\color{tableborder}\vrule}}
            \rowcolor{errorcolor}\textbf{LLaVA 1.5 (cmp)}: & \textbf{This is a collage of various activities} \\
            \rowcolor{errorcolor}\textbf{LLaVA 1.6 (cmp)}: & \textbf{Collage} \\
            \rowcolor{errorcolor}\textbf{Moondream2 (cmp)}: & \textbf{Collage} \\
            \rowcolor{answercolor}\textbf{Mantis (int)}: & \textbf{a person riding a horse} \\
            \rowcolor{errorcolor}\textbf{GPT-4V (cmp)}: & \textbf{surfing} \\
            \rowcolor{answercolor}\textbf{GPT-4V (int)}: & \textbf{Riding a horse} \\
            \rowcolor{errorcolor}\textbf{Gemini 1.0 (cmp)}: & \textbf{surfing} \\
            \rowcolor{errorcolor}\textbf{Gemini 1.0 (int)}: & \textbf{surfing} \\
            \rowcolor{answercolor}\textbf{Gemini 1.5 (cmp)}: & \textbf{horseback riding} \\
            \rowcolor{errorcolor}\textbf{Gemini 1.5 (int)}: & \textbf{Surfing} \\
            \rowcolor{errorcolor}\textbf{PaliGemma (cmp)}: & \textbf{collage} \\
        \end{tabularx} \\ \\
        \hline
    \end{tabular}
    \begin{tabular}{|>{\centering\arraybackslash}m{0.9\linewidth}|}
        \rowcolor{bluehighlight}\textbf{Visual Context Length ($k=25$)} \label{acs_okvqa_k25} \\
        \addcontentsline{toc}{subsection}{25-Image Context}
        \begin{tabularx}{0.913\textwidth}{!{\color{tableborder}\vrule} X c !{\color{tableborder}\vrule}}
            \rowcolor{errorcolor}\textbf{LLaVA 1.5 (cmp)}: & \textbf{This is a collage of photos, showcasing various activities.} \\
            \rowcolor{errorcolor}\textbf{LLaVA 1.6 (cmp)}: & \textbf{Collage} \\
            \rowcolor{errorcolor}\textbf{Moondream2 (cmp)}: & \textbf{Collage} \\
            \rowcolor{answercolor}\textbf{Mantis (int)}: & \textbf{Horseback riding} \\
            \rowcolor{errorcolor}\textbf{GPT-4V (cmp)}: & \textbf{skiiing} \\
            \rowcolor{errorcolor}\textbf{GPT-4V (int)}: & \textbf{playing catch with a frisbee} \\
            \rowcolor{errorcolor}\textbf{Gemini 1.0 (cmp)}: & \textbf{This is a photo collage} \\
            \rowcolor{errorcolor}\textbf{Gemini 1.0 (int)}: & \textbf{This is a photo collage} \\
            \rowcolor{answercolor}\textbf{Gemini 1.5 (cmp)}: & \textbf{horseback riding} \\
            \rowcolor{errorcolor}\textbf{Gemini 1.5 (int)}: & \textbf{frisbee} \\
            \rowcolor{errorcolor}\textbf{PaliGemma (cmp)}: & \textbf{skiing} \\
        \end{tabularx} \\ \\
        \hline
    \end{tabular}
    \begin{tabular}{|>{\centering\arraybackslash}m{0.9\linewidth}|}
        \rowcolor{bluehighlight}\textbf{Visual Context Length ($k=36$)} \label{acs_okvqa_k36}\\
        \addcontentsline{toc}{subsection}{36-Image Context}
        \begin{tabularx}{0.913\textwidth}{!{\color{tableborder}\vrule} X c !{\color{tableborder}\vrule}}
            \rowcolor{errorcolor}\textbf{LLaVA 1.5 (cmp)}: & \textbf{a man eating a piece of cake} \\
            \rowcolor{errorcolor}\textbf{LLaVA 1.6 (cmp)}: & \textbf{motorcycle} \\
            \rowcolor{errorcolor}\textbf{Moondream2 (cmp)}: & \textbf{colleage} \\
            \rowcolor{errorcolor}\textbf{Mantis (int)}: & \textbf{small garden with a shed} \\
            \rowcolor{errorcolor}\textbf{GPT-4V (cmp)}: & \textbf{surfing} \\
            \rowcolor{errorcolor}\textbf{GPT-4V (int)}: & \textbf{Unable to provide the activities} \\
            \rowcolor{errorcolor}\textbf{Gemini 1.0 (cmp)}: & \textbf{This is a game of Where's Waldo.} \\
            \rowcolor{errorcolor}\textbf{Gemini 1.0 (int)}: & \textbf{This is a game of Where's Waldo.} \\
            \rowcolor{errorcolor}\textbf{Gemini 1.5 (cmp)}: & \textbf{Surfing} \\
            \rowcolor{errorcolor}\textbf{Gemini 1.5 (int)}: & \textbf{life} \\
            \rowcolor{errorcolor}\textbf{PaliGemma (cmp)}: & \textbf{collage} \\
        \end{tabularx} \\
        \hline
    \end{tabular}
\end{center}
\newpage
\begin{center}
    \arrayrulecolor{tableborder}
    \begin{tabular}{|>{\centering\arraybackslash}m{0.9\linewidth}|}
        \hline
        \rowcolor[gray]{0.9} 
        \textbf{\large Example 2 -- MMStar (SeedBench)}
        \addcontentsline{toc}{section}{MMStar (SeedBench)}
    \end{tabular}
    \begin{tabular}{|>{\raggedright\arraybackslash}m{0.9\linewidth}|}
        \hline
        \rowcolor{questioncolor}\textbf{Question:} What is the color of the socks detected in the attribute detection of the \texttt{<image>}? \\
        \rowcolor{questioncolor}\begin{enumerate}[label=(\Alph*), itemsep=1.2pt, parsep=0pt]
                \item Black
                \item Red
                \item White
                \item Striped
            \end{enumerate}\\
        \hline
        \begin{center}
            \fbox{\includegraphics[width=0.2\linewidth]{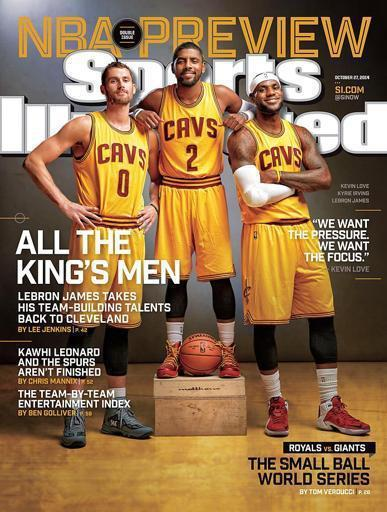}} \\
            \texttt{<image>}
            \label{mmstar_sb}
        \end{center} \\
        \hline
    \end{tabular}
    \begin{tabular}{|>{\centering\arraybackslash}m{0.9\linewidth}|}
        \rowcolor{bluehighlight}\textbf{Visual Context Length ($k=1$)} \label{mmstar_sb_k1} \\
        \addcontentsline{toc}{subsection}{1-Image Context}
        \begin{tabularx}{0.913\textwidth}{!{\color{tableborder}\vrule} X c !{\color{tableborder}\vrule}}
            \rowcolor{answercolor}\textbf{LLaVA 1.5 (cmp)}: & \textbf{C} \\
            \rowcolor{errorcolor}\textbf{LLaVA 1.6 (cmp)}: & \textbf{B} \\
            \rowcolor{errorcolor}\textbf{Moondream2 (cmp)}: & \textbf{A} \\
            \rowcolor{errorcolor}\textbf{Mantis (int)}: & \textbf{B} \\
            \rowcolor{answercolor}\textbf{GPT-4V (cmp)}: & \textbf{C} \\
            \rowcolor{answercolor}\textbf{GPT-4V (int)}: & \textbf{C} \\
            \rowcolor{errorcolor}\textbf{Gemini 1.0 (cmp)}: & \textbf{B} \\
            \rowcolor{errorcolor}\textbf{Gemini 1.0 (int)}: & \textbf{B} \\
            \rowcolor{errorcolor}\textbf{Gemini 1.5 (cmp)}: & \textbf{A} \\
            \rowcolor{errorcolor}\textbf{Gemini 1.5 (int)}: & \textbf{A} \\
            \rowcolor{answercolor}\textbf{PaliGemma (cmp)}: & \textbf{C} \\
        \end{tabularx} \\ \\
        \hline
    \end{tabular}
    \begin{tabular}{|>{\centering\arraybackslash}m{0.9\linewidth}|}
        \rowcolor{bluehighlight}\textbf{Visual Context Length ($k=4$)} 
        \label{mmstar_sb_k4}\\
        \addcontentsline{toc}{subsection}{4-Image Context}
        \begin{tabularx}{0.913\textwidth}{!{\color{tableborder}\vrule} X c !{\color{tableborder}\vrule}}
            \rowcolor{errorcolor}\textbf{LLaVA 1.5 (cmp)}: & \textbf{A} \\
            \rowcolor{errorcolor}\textbf{LLaVA 1.6 (cmp)}: & \textbf{B} \\
            \rowcolor{errorcolor}\textbf{Moondream2 (cmp)}: & \textbf{A} \\
            \rowcolor{errorcolor}\textbf{Mantis (int)}: & \textbf{B} \\
            \rowcolor{answercolor}\textbf{GPT-4V (cmp)}: & \textbf{C} \\
            \rowcolor{answercolor}\textbf{GPT-4V (int)}: & \textbf{C} \\
            \rowcolor{errorcolor}\textbf{Gemini 1.0 (cmp)}: & \textbf{D} \\
            \rowcolor{answercolor}\textbf{Gemini 1.0 (int)}: & \textbf{C} \\
            \rowcolor{answercolor}\textbf{Gemini 1.5 (cmp)}: & \textbf{C} \\
            \rowcolor{answercolor}\textbf{Gemini 1.5 (int)}: & \textbf{C} \\
            \rowcolor{errorcolor}\textbf{PaliGemma (cmp)}: & \textbf{A} \\
        \end{tabularx} \\
        \hline
    \end{tabular}
    \begin{tabular}{|>{\centering\arraybackslash}m{0.9\linewidth}|}
        \rowcolor{bluehighlight}\textbf{Visual Context Length ($k=9$)}
        \label{mmstar_sb_k9}\\
        \addcontentsline{toc}{subsection}{9-Image Context}
        \begin{tabularx}{0.913\textwidth}{!{\color{tableborder}\vrule} X c !{\color{tableborder}\vrule}}
            \rowcolor{errorcolor}\textbf{LLaVA 1.5 (cmp)}: & \textbf{A} \\
            \rowcolor{errorcolor}\textbf{LLaVA 1.6 (cmp)}: & \textbf{B} \\
            \rowcolor{errorcolor}\textbf{Moondream2 (cmp)}: & \textbf{A} \\
            \rowcolor{errorcolor}\textbf{Mantis (int)}: & \textbf{A: Black, B: Red, C: White} \\
            \rowcolor{errorcolor}\textbf{GPT-4V (cmp)}: & \textbf{There are no visible socks in the image.} \\
            \rowcolor{errorcolor}\textbf{GPT-4V (int)}: & \textbf{None of the images provided.} \\
            \rowcolor{answercolor}\textbf{Gemini 1.0 (cmp)}: & \textbf{C} \\
            \rowcolor{answercolor}\textbf{Gemini 1.0 (int)}: & \textbf{C} \\
            \rowcolor{errorcolor}\textbf{Gemini 1.5 (cmp)}: & \textbf{Blue} \\
            \rowcolor{errorcolor}\textbf{Gemini 1.5 (int)}: & \textbf{There are no socks in the image.} \\
            \rowcolor{errorcolor}\textbf{PaliGemma (cmp)}: & \textbf{A} \\
        \end{tabularx} \\ \\
        \hline
    \end{tabular}
    \begin{tabular}{|>{\centering\arraybackslash}m{0.9\linewidth}|}
        \rowcolor{bluehighlight}\textbf{Visual Context Length ($k=16$)}
        \label{mmstar_sb_k16}\\
        \addcontentsline{toc}{subsection}{16-Image Context}
        \begin{tabularx}{0.913\textwidth}{!{\color{tableborder}\vrule} X c !{\color{tableborder}\vrule}}
            \rowcolor{errorcolor}\textbf{LLaVA 1.5 (cmp)} & \textbf{A} \\
            \rowcolor{errorcolor}\textbf{LLaVA 1.6 (cmp)} & \textbf{B} \\
            \rowcolor{errorcolor}\textbf{Moondream2 (cmp)} & \textbf{A} \\
            \rowcolor{errorcolor}\textbf{Mantis (int)} & \textbf{A: Black, B: Red, C: White, D: Striped} \\
            \rowcolor{errorcolor}\textbf{GPT-4V (cmp)} & \textbf{There are no socks visible in the image} \\
            \rowcolor{errorcolor}\textbf{GPT-4V (int)} & \textbf{None of the above.} \\
            \rowcolor{answercolor}\textbf{Gemini 1.0 (cmp)} & \textbf{C} \\
            \rowcolor{answercolor}\textbf{Gemini 1.0 (int)} & \textbf{C} \\
            \rowcolor{answercolor}\textbf{Gemini 1.5 (cmp)} & \textbf{C} \\
            \rowcolor{errorcolor}\textbf{Gemini 1.5 (int)} & \textbf{There are no socks in the image.} \\
            \rowcolor{answercolor}\textbf{PaliGemma (cmp)} & \textbf{C} \\
        \end{tabularx} \\ \\
        \hline
    \end{tabular}
    \begin{tabular}{|>{\centering\arraybackslash}m{0.9\linewidth}|}
        \rowcolor{bluehighlight}\textbf{Visual Context Length ($k=25$)}
        \label{mmstar_sb_k25}\\
        \addcontentsline{toc}{subsection}{25-Image Context}
        \begin{tabularx}{0.913\textwidth}{!{\color{tableborder}\vrule} X c !{\color{tableborder}\vrule}}
            \rowcolor{answercolor}\textbf{LLaVA 1.5 (cmp)} & \textbf{C} \\
            \rowcolor{errorcolor}\textbf{LLaVA 1.6 (cmp)} & \textbf{B} \\
            \rowcolor{errorcolor}\textbf{Moondream2 (cmp)} & \textbf{A} \\
            \rowcolor{errorcolor}\textbf{Mantis (int)} & \textbf{B} \\ \rowcolor{answercolor}\textbf{GPT-4V (cmp)} & \textbf{C} \\
            \rowcolor{errorcolor}\textbf{GPT-4V (int)} & \textbf{Option C is not provided} \\
            \rowcolor{answercolor}\textbf{Gemini 1.0 (cmp)} & \textbf{C} \\
            \rowcolor{answercolor}\textbf{Gemini 1.0 (int)} & \textbf{C} \\
            \rowcolor{errorcolor}\textbf{Gemini 1.5 (cmp)} & \textbf{There are no socks in the image.} \\
            \rowcolor{errorcolor}\textbf{Gemini 1.5 (int)} & \textbf{There are no socks in the image.} \\
            \rowcolor{errorcolor}\textbf{PaliGemma (cmp)} & \textbf{B} \\
        \end{tabularx} \\ \\
        \hline
    \end{tabular}
    \begin{tabular}{|>{\centering\arraybackslash}m{0.9\linewidth}|}
        \rowcolor{bluehighlight}\textbf{Visual Context Length ($k=36$)}
        \label{mmstar_sb_k36}\\
        \addcontentsline{toc}{subsection}{36-Image Context}
        \begin{tabularx}{0.913\textwidth}{!{\color{tableborder}\vrule} X c !{\color{tableborder}\vrule}}
            \rowcolor{answercolor}\textbf{LLaVA 1.5 (cmp)} & \textbf{C} \\
            \rowcolor{errorcolor}\textbf{LLaVA 1.6 (cmp)} & \textbf{B} \\
            \rowcolor{errorcolor}\textbf{Moondream2 (cmp)} & \textbf{B} \\
            \rowcolor{answercolor}\textbf{Mantis (int)} & \textbf{C} \\ \rowcolor{errorcolor}\textbf{GPT-4V (cmp)} & \textbf{I am unable to provide assistance with this request} \\
            \rowcolor{errorcolor}\textbf{GPT-4V (int)} & \textbf{None of these} \\
            \rowcolor{errorcolor}\textbf{Gemini 1.0 (cmp)} & \textbf{A} \\
            \rowcolor{answercolor}\textbf{Gemini 1.0 (int)} & \textbf{C} \\
            \rowcolor{errorcolor}\textbf{Gemini 1.5 (cmp)} & \textbf{There are no socks in the image.} \\
            \rowcolor{errorcolor}\textbf{Gemini 1.5 (int)} & \textbf{There are no socks in the image.} \\
            \rowcolor{errorcolor}\textbf{PaliGemma (cmp)} & \textbf{B} \\
        \end{tabularx} \\
        \hline
    \end{tabular}
\end{center}
\newpage
\newpage
\begin{center}
    \arrayrulecolor{tableborder}
    \begin{tabular}{|>{\centering\arraybackslash}m{0.9\linewidth}|}
        \hline
        \rowcolor[gray]{0.9}\textbf{\large Example 3 -- MMStar (MMBench)}
        \addcontentsline{toc}{section}{MMStar (MMBench)}
    \end{tabular}
    \begin{tabular}{|>{\raggedright\arraybackslash}m{0.9\linewidth}|}
        \hline
        \rowcolor{questioncolor}\textbf{Question:} Which is the main topic of the \texttt{<image>}? \\
        \rowcolor{questioncolor}\begin{enumerate}[label=(\Alph*), itemsep=1.2pt, parsep=0pt]
                \item A woman surfing
                \item A man skating
                \item A man surfing
                \item A woman skating
            \end{enumerate}\\
        \hline
        \begin{center}
            \fbox{\includegraphics[width=0.2\linewidth]{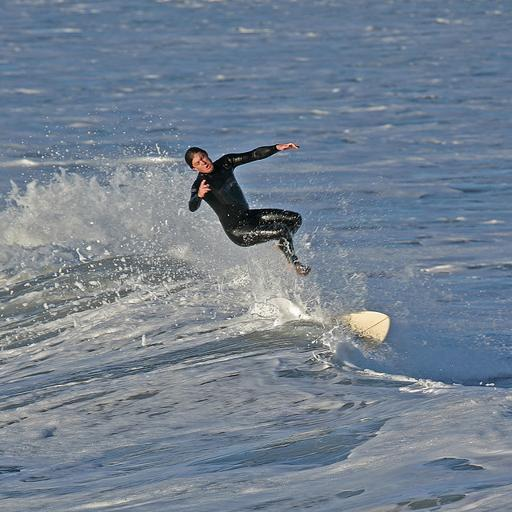}} \\
            \texttt{<image>}
            \label{mmstar_mb}
        \end{center} \\
        \hline
    \end{tabular}
    \begin{tabular}{|>{\centering\arraybackslash}m{0.9\linewidth}|}
        \rowcolor{bluehighlight}\textbf{Visual Context Length ($k=1$)} \\
        \addcontentsline{toc}{subsection}{1-Image Context}
        \begin{tabularx}{0.913\textwidth}{!{\color{tableborder}\vrule} X c !{\color{tableborder}\vrule}}
            \rowcolor{answercolor}\textbf{LLaVA 1.5 (cmp)}: & \textbf{C} \\
            \rowcolor{answercolor}\textbf{LLaVA 1.6 (cmp)}: & \textbf{C} \\
            \rowcolor{errorcolor}\textbf{Moondream2 (cmp)}: & \textbf{A} \\
            \rowcolor{answercolor}\textbf{Mantis (int)}: & \textbf{C} \\
            \rowcolor{answercolor}\textbf{GPT-4V (cmp)}: & \textbf{C} \\
            \rowcolor{answercolor}\textbf{GPT-4V (int)}: & \textbf{C} \\
            \rowcolor{answercolor}\textbf{Gemini 1.0 (cmp)}: & \textbf{C} \\
            \rowcolor{answercolor}\textbf{Gemini 1.0 (int)}: & \textbf{C} \\
            \rowcolor{answercolor}\textbf{Gemini 1.5 (cmp)}: & \textbf{C} \\
            \rowcolor{answercolor}\textbf{Gemini 1.5 (int)}: & \textbf{C} \\
            \rowcolor{answercolor}\textbf{PaliGemma (cmp)}: & \textbf{C} \\
        \end{tabularx} \\ \\
        \hline
    \end{tabular}
    \begin{tabular}{|>{\centering\arraybackslash}m{0.9\linewidth}|}
        \rowcolor{bluehighlight}\textbf{Visual Context Length ($k=4$)} \\
        \addcontentsline{toc}{subsection}{4-Image Context}
        \begin{tabularx}{0.913\textwidth}{!{\color{tableborder}\vrule} X c !{\color{tableborder}\vrule}}
            \rowcolor{answercolor}\textbf{LLaVA 1.5 (cmp)}: & \textbf{C} \\
            \rowcolor{answercolor}\textbf{LLaVA 1.6 (cmp)}: & \textbf{C} \\
            \rowcolor{errorcolor}\textbf{Moondream2 (cmp)}: & \textbf{A} \\
            \rowcolor{answercolor}\textbf{Mantis (int)}: & \textbf{C} \\
            \rowcolor{answercolor}\textbf{GPT-4V (cmp)}: & \textbf{C} \\
            \rowcolor{answercolor}\textbf{GPT-4V (int)}: & \textbf{C} \\
            \rowcolor{answercolor}\textbf{Gemini 1.0 (cmp)}: & \textbf{C} \\
            \rowcolor{answercolor}\textbf{Gemini 1.0 (int)}: & \textbf{C} \\
            \rowcolor{answercolor}\textbf{Gemini 1.5 (cmp)}: & \textbf{C} \\
            \rowcolor{answercolor}\textbf{Gemini 1.5 (int)}: & \textbf{C} \\
            \rowcolor{answercolor}\textbf{PaliGemma (cmp)}: & \textbf{C} \\
        \end{tabularx} \\
        \hline
    \end{tabular}
    \begin{tabular}{|>{\centering\arraybackslash}m{0.9\linewidth}|}
        \rowcolor{bluehighlight}\textbf{Visual Context Length ($k=9$)} \\
        \addcontentsline{toc}{subsection}{9-Image Context}
        \begin{tabularx}{0.913\textwidth}{!{\color{tableborder}\vrule} X c !{\color{tableborder}\vrule}}
            \rowcolor{answercolor}\textbf{LLaVA 1.5 (cmp)}: & \textbf{C} \\
            \rowcolor{answercolor}\textbf{LLaVA 1.6 (cmp)}: & \textbf{C} \\
            \rowcolor{errorcolor}\textbf{Moondream2 (cmp)}: & \textbf{A} \\
            \rowcolor{errorcolor}\textbf{Mantis (int)}: & \textbf{C: A man surfing, D: A woman skiting} \\
            \rowcolor{answercolor}\textbf{GPT-4V (cmp)}: & \textbf{C} \\
            \rowcolor{errorcolor}\textbf{GPT-4V (int)}: & \textbf{-} \\
            \rowcolor{errorcolor}\textbf{Gemini 1.0 (cmp)}: & \textbf{A} \\
            \rowcolor{answercolor}\textbf{Gemini 1.0 (int)}: & \textbf{C} \\
            \rowcolor{answercolor}\textbf{Gemini 1.5 (cmp)}: & \textbf{C} \\
            \rowcolor{answercolor}\textbf{Gemini 1.5 (int)}: & \textbf{C} \\
            \rowcolor{answercolor}\textbf{PaliGemma (cmp)}: & \textbf{C} \\
        \end{tabularx} \\ \\
        \hline
    \end{tabular}
    \begin{tabular}{|>{\centering\arraybackslash}m{0.9\linewidth}|}
        \rowcolor{bluehighlight}\textbf{Visual Context Length ($k=16$)} \\
        \addcontentsline{toc}{subsection}{16-Image Context}
        \begin{tabularx}{0.913\textwidth}{!{\color{tableborder}\vrule} X c !{\color{tableborder}\vrule}}
            \rowcolor{answercolor}\textbf{LLaVA 1.5 (cmp)}: & \textbf{C} \\
            \rowcolor{answercolor}\textbf{LLaVA 1.6 (cmp)}: & \textbf{C} \\
            \rowcolor{errorcolor}\textbf{Moondream2 (cmp)}: & \textbf{A} \\
            \rowcolor{errorcolor}\textbf{Mantis (int)}: & \textbf{C: A man surfing, D: A woman skiting} \\
            \rowcolor{answercolor}\textbf{GPT-4V (cmp)}: & \textbf{C} \\
            \rowcolor{errorcolor}\textbf{GPT-4V (int)}: & \textbf{-} \\
            \rowcolor{errorcolor}\textbf{Gemini 1.0 (cmp)}: & \textbf{A} \\
            \rowcolor{errorcolor}\textbf{Gemini 1.0 (int)}: & \textbf{A} \\
            \rowcolor{answercolor}\textbf{Gemini 1.5 (cmp)}: & \textbf{C} \\
            \rowcolor{answercolor}\textbf{Gemini 1.5 (int)}: & \textbf{C} \\
            \rowcolor{answercolor}\textbf{PaliGemma (cmp)}: & \textbf{C} \\
        \end{tabularx} \\ \\
        \hline
    \end{tabular}
    \begin{tabular}{|>{\centering\arraybackslash}m{0.9\linewidth}|}
        \rowcolor{bluehighlight}\textbf{Visual Context Length ($k=25$)} \\
        \addcontentsline{toc}{subsection}{25-Image Context}
        \begin{tabularx}{0.913\textwidth}{!{\color{tableborder}\vrule} X c !{\color{tableborder}\vrule}}
            \rowcolor{answercolor}\textbf{LLaVA 1.5 (cmp)}: & \textbf{C} \\
            \rowcolor{errorcolor}\textbf{LLaVA 1.6 (cmp)}: & \textbf{A} \\
            \rowcolor{errorcolor}\textbf{Moondream2 (cmp)}: & \textbf{A} \\
            \rowcolor{errorcolor}\textbf{Mantis (int)}: & \textbf{B} \\
            \rowcolor{errorcolor}\textbf{GPT-4V (cmp)}: & \textbf{-} \\
            \rowcolor{errorcolor}\textbf{GPT-4V (int)}: & \textbf{-} \\
            \rowcolor{errorcolor}\textbf{Gemini 1.0 (cmp)}: & \textbf{A} \\
            \rowcolor{errorcolor}\textbf{Gemini 1.0 (int)}: & \textbf{A} \\
            \rowcolor{answercolor}\textbf{Gemini 1.5 (cmp)}: & \textbf{C} \\
            \rowcolor{errorcolor}\textbf{Gemini 1.5 (int)}: & \textbf{B} \\
            \rowcolor{errorcolor}\textbf{PaliGemma (cmp)}: & \textbf{D} \\
        \end{tabularx} \\ \\
        \hline
    \end{tabular}
    \begin{tabular}{|>{\centering\arraybackslash}m{0.9\linewidth}|}
        \rowcolor{bluehighlight}\textbf{Visual Context Length ($k=36$)} \\
        \addcontentsline{toc}{subsection}{36-Image Context}
        \begin{tabularx}{0.913\textwidth}{!{\color{tableborder}\vrule} X c !{\color{tableborder}\vrule}}
            \rowcolor{answercolor}\textbf{LLaVA 1.5 (cmp)}: & \textbf{C} \\
            \rowcolor{errorcolor}\textbf{LLaVA 1.6 (cmp)}: & \textbf{A} \\
            \rowcolor{errorcolor}\textbf{Moondream2 (cmp)}: & \textbf{A} \\
            \rowcolor{errorcolor}\textbf{Mantis (int)}: & \textbf{D} \\
            \rowcolor{answercolor}\textbf{GPT-4V (cmp)}: & \textbf{C} \\
            \rowcolor{errorcolor}\textbf{GPT-4V (int)}: & \textbf{-} \\
            \rowcolor{errorcolor}\textbf{Gemini 1.0 (cmp)}: & \textbf{A} \\
            \rowcolor{errorcolor}\textbf{Gemini 1.0 (int)}: & \textbf{A} \\
            \rowcolor{answercolor}\textbf{Gemini 1.5 (cmp)}: & \textbf{C} \\
            \rowcolor{answercolor}\textbf{Gemini 1.5 (int)}: & \textbf{C} \\
            \rowcolor{answercolor}\textbf{PaliGemma (cmp)}: & \textbf{C} \\
        \end{tabularx} \\
        \hline
    \end{tabular}
\end{center}

\newpage

\begin{center}
    \arrayrulecolor{tableborder}
    \begin{tabular}{|>{\centering\arraybackslash}m{0.9\linewidth}|}
        \hline
        \rowcolor[gray]{0.9}\textbf{\large Example 4 -- MMStar (MathVista)}
        \addcontentsline{toc}{section}{MMStar (MathVista)}
    \end{tabular}
    \begin{tabular}{|>{\raggedright\arraybackslash}m{0.9\linewidth}|}
        \hline
        \rowcolor{questioncolor}\textbf{Question:} What is the age gap between these two people in \texttt{<image>}? (Unit: years) \\
        \rowcolor{questioncolor}\begin{enumerate}[label=(\Alph*), itemsep=1.2pt, parsep=0pt]
                \item 4
                \item 3
                \item 2
                \item 1
            \end{enumerate}\\
        \hline
        \begin{center}
            \fbox{\includegraphics[width=0.4\linewidth]{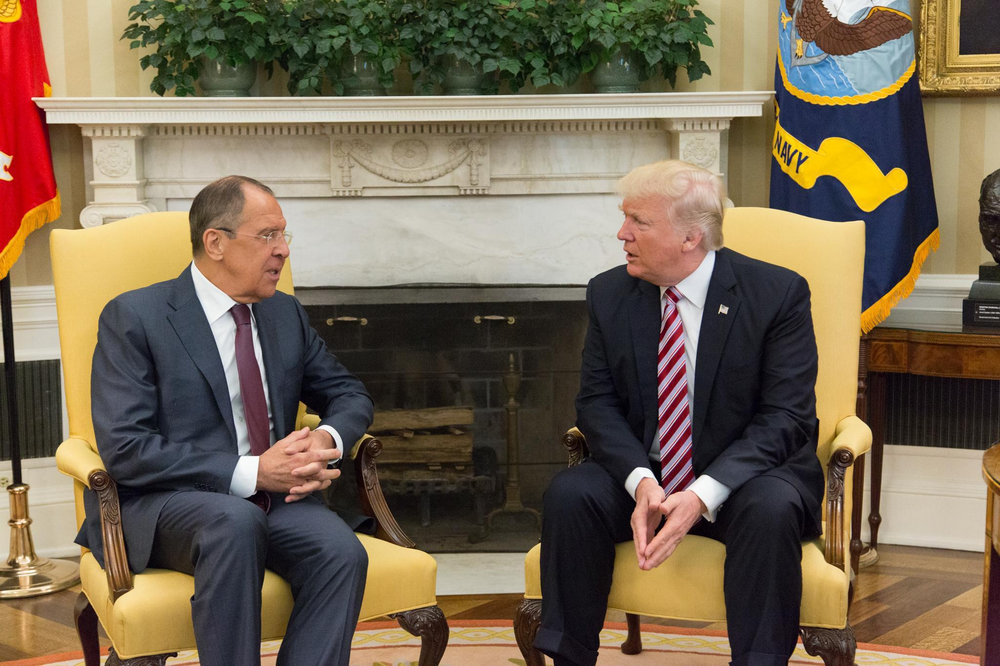}} \\
            \texttt{<image>}
            \label{mmstar_mv}
        \end{center} \\
        \hline
    \end{tabular}
    \begin{tabular}{|>{\centering\arraybackslash}m{0.9\linewidth}|}
        \rowcolor{bluehighlight}\textbf{Visual Context Length ($k=1$)} \\
        \addcontentsline{toc}{subsection}{1-Image Context}
        \begin{tabularx}{0.913\textwidth}{!{\color{tableborder}\vrule} X c !{\color{tableborder}\vrule}}
            \rowcolor{errorcolor}\textbf{LLaVA 1.5 (cmp)}: & \textbf{E} \\
            \rowcolor{answercolor}\textbf{LLaVA 1.6 (cmp)}: & \textbf{A} \\
            \rowcolor{answercolor}\textbf{Moondream2 (cmp)}: & \textbf{A} \\
            \rowcolor{answercolor}\textbf{Mantis (int)}: & \textbf{A} \\
            \rowcolor{errorcolor}\textbf{GPT-4V (cmp)}: & \textbf{I'm sorry, but I cannot provide information} \\
            \rowcolor{errorcolor}\textbf{GPT-4V (int)}: & \textbf{I'm sorry, but I cannot provide information} \\
            \rowcolor{answercolor}\textbf{Gemini 1.0 (cmp)}: & \textbf{A} \\
            \rowcolor{answercolor}\textbf{Gemini 1.0 (int)}: & \textbf{A} \\
            \rowcolor{answercolor}\textbf{Gemini 1.5 (cmp)}: & \textbf{A} \\
            \rowcolor{answercolor}\textbf{Gemini 1.5 (int)}: & \textbf{A} \\
            \rowcolor{answercolor}\textbf{PaliGemma (cmp)}: & \textbf{A} \\
        \end{tabularx} \\ \\
        \hline
    \end{tabular}
    \begin{tabular}{|>{\centering\arraybackslash}m{0.9\linewidth}|}
        \rowcolor{bluehighlight}\textbf{Visual Context Length ($k=4$)} \\
        \addcontentsline{toc}{subsection}{4-Image Context}
        \begin{tabularx}{0.913\textwidth}{!{\color{tableborder}\vrule} X c !{\color{tableborder}\vrule}}
            \rowcolor{errorcolor}\textbf{LLaVA 1.5 (cmp)}: & \textbf{D} \\
            \rowcolor{answercolor}\textbf{LLaVA 1.6 (cmp)}: & \textbf{A} \\
            \rowcolor{errorcolor}\textbf{Moondream2 (cmp)}: & \textbf{B} \\
            \rowcolor{errorcolor}\textbf{Mantis (int)}: & \textbf{B} \\
            \rowcolor{errorcolor}\textbf{GPT-4V (cmp)}: & \textbf{I'm sorry, but I cannot provide information} \\
            \rowcolor{errorcolor}\textbf{GPT-4V (int)}: & \textbf{I'm sorry, but I cannot provide information} \\
            \rowcolor{errorcolor}\textbf{Gemini 1.0 (cmp)}: & \textbf{B} \\
            \rowcolor{errorcolor}\textbf{Gemini 1.0 (int)}: & \textbf{B} \\
            \rowcolor{errorcolor}\textbf{Gemini 1.5 (cmp)}: & \textbf{D} \\
            \rowcolor{errorcolor}\textbf{Gemini 1.5 (int)}: & \textbf{C} \\
            \rowcolor{errorcolor}\textbf{PaliGemma (cmp)}: & \textbf{B} \\
        \end{tabularx} \\
        \hline
    \end{tabular}
    \begin{tabular}{|>{\centering\arraybackslash}m{0.9\linewidth}|}
        \rowcolor{bluehighlight}\textbf{Visual Context Length ($k=9$)} \\
        \addcontentsline{toc}{subsection}{9-Image Context}
        \begin{tabularx}{0.913\textwidth}{!{\color{tableborder}\vrule} X c !{\color{tableborder}\vrule}}
            \rowcolor{errorcolor}\textbf{LLaVA 1.5 (cmp)}: & \textbf{E} \\
            \rowcolor{answercolor}\textbf{LLaVA 1.6 (cmp)}: & \textbf{A} \\
            \rowcolor{errorcolor}\textbf{Moondream2 (cmp)}: & \textbf{B} \\
            \rowcolor{errorcolor}\textbf{Mantis (int)}: & \textbf{B} \\
            \rowcolor{errorcolor}\textbf{GPT-4V (cmp)}: & \textbf{I'm sorry, but I cannot provide information} \\
            \rowcolor{errorcolor}\textbf{GPT-4V (int)}: & \textbf{I'm sorry, but I cannot provide information} \\
            \rowcolor{answercolor}\textbf{Gemini 1.0 (cmp)}: & \textbf{A} \\
            \rowcolor{errorcolor}\textbf{Gemini 1.0 (int)}: & \textbf{D} \\
            \rowcolor{errorcolor}\textbf{Gemini 1.5 (cmp)}: & \textbf{C} \\
            \rowcolor{errorcolor}\textbf{Gemini 1.5 (int)}: & \textbf{C} \\
            \rowcolor{errorcolor}\textbf{PaliGemma (cmp)}: & \textbf{B} \\
        \end{tabularx} \\ \\
        \hline
    \end{tabular}
    \begin{tabular}{|>{\centering\arraybackslash}m{0.9\linewidth}|}
        \rowcolor{bluehighlight}\textbf{Visual Context Length ($k=16$)} \\
        \addcontentsline{toc}{subsection}{16-Image Context}
        \begin{tabularx}{0.913\textwidth}{!{\color{tableborder}\vrule} X c !{\color{tableborder}\vrule}}
            \rowcolor{errorcolor}\textbf{LLaVA 1.5 (cmp)}: & \textbf{C} \\
            \rowcolor{answercolor}\textbf{LLaVA 1.6 (cmp)}: & \textbf{A} \\
            \rowcolor{errorcolor}\textbf{Moondream2 (cmp)}: & \textbf{B} \\
            \rowcolor{answercolor}\textbf{Mantis (int)}: & \textbf{A} \\
            \rowcolor{errorcolor}\textbf{GPT-4V (cmp)}: & \textbf{I'm sorry, but I cannot provide information} \\
            \rowcolor{errorcolor}\textbf{GPT-4V (int)}: & \textbf{I'm sorry, but I cannot provide information} \\
            \rowcolor{errorcolor}\textbf{Gemini 1.0 (cmp)}: & \textbf{B} \\
            \rowcolor{errorcolor}\textbf{Gemini 1.0 (int)}: & \textbf{B} \\
            \rowcolor{errorcolor}\textbf{Gemini 1.5 (cmp)}: & \textbf{C} \\
            \rowcolor{errorcolor}\textbf{Gemini 1.5 (int)}: & \textbf{D} \\
            \rowcolor{errorcolor}\textbf{PaliGemma (cmp)}: & \textbf{B} \\
        \end{tabularx} \\ \\
        \hline
    \end{tabular}
    \begin{tabular}{|>{\centering\arraybackslash}m{0.9\linewidth}|}
        \rowcolor{bluehighlight}\textbf{Visual Context Length ($k=25$)} \\
        \addcontentsline{toc}{subsection}{25-Image Context}
        \begin{tabularx}{0.913\textwidth}{!{\color{tableborder}\vrule} X c !{\color{tableborder}\vrule}}
            \rowcolor{errorcolor}\textbf{LLaVA 1.5 (cmp)}: & \textbf{E} \\
            \rowcolor{answercolor}\textbf{LLaVA 1.6 (cmp)}: & \textbf{A} \\
            \rowcolor{errorcolor}\textbf{Moondream2 (cmp)}: & \textbf{B} \\
            \rowcolor{errorcolor}\textbf{Mantis (int)}: & \textbf{B} \\
            \rowcolor{errorcolor}\textbf{GPT-4V (cmp)}: & \textbf{I'm sorry, but I cannot provide information} \\
            \rowcolor{errorcolor}\textbf{GPT-4V (int)}: & \textbf{-} \\
            \rowcolor{errorcolor}\textbf{Gemini 1.0 (cmp)}: & \textbf{D} \\
            \rowcolor{errorcolor}\textbf{Gemini 1.0 (int)}: & \textbf{D} \\
            \rowcolor{errorcolor}\textbf{Gemini 1.5 (cmp)}: & \textbf{C} \\
            \rowcolor{errorcolor}\textbf{Gemini 1.5 (int)}: & \textbf{C} \\
            \rowcolor{answercolor}\textbf{PaliGemma (cmp)}: & \textbf{A} \\
        \end{tabularx} \\ \\
        \hline
    \end{tabular}
    \begin{tabular}{|>{\centering\arraybackslash}m{0.9\linewidth}|}
        \rowcolor{bluehighlight}\textbf{Visual Context Length ($k=36$)} \\
        \addcontentsline{toc}{subsection}{36-Image Context}
        \begin{tabularx}{0.913\textwidth}{!{\color{tableborder}\vrule} X c !{\color{tableborder}\vrule}}
            \rowcolor{errorcolor}\textbf{LLaVA 1.5 (cmp)}: & \textbf{D} \\
            \rowcolor{answercolor}\textbf{LLaVA 1.6 (cmp)}: & \textbf{A} \\
            \rowcolor{errorcolor}\textbf{Moondream2 (cmp)}: & \textbf{B} \\
            \rowcolor{errorcolor}\textbf{Mantis (int)}: & \textbf{B} \\
            \rowcolor{errorcolor}\textbf{GPT-4V (cmp)}: & \textbf{-} \\
            \rowcolor{errorcolor}\textbf{GPT-4V (int)}: & \textbf{Impossible to determine} \\
            \rowcolor{answercolor}\textbf{Gemini 1.0 (cmp)}: & \textbf{A} \\
            \rowcolor{answercolor}\textbf{Gemini 1.0 (int)}: & \textbf{A} \\
            \rowcolor{errorcolor}\textbf{Gemini 1.5 (cmp)}: & \textbf{D} \\
            \rowcolor{errorcolor}\textbf{Gemini 1.5 (int)}: & \textbf{C} \\
            \rowcolor{errorcolor}\textbf{PaliGemma (cmp)}: & \textbf{B} \\
        \end{tabularx} \\
        \hline
    \end{tabular}
\end{center}

\end{document}